\pdfoutput=1
\documentclass[11pt]{article}

\usepackage[]{EMNLP2023}

\usepackage{times}
\usepackage{amsmath}
\usepackage{latexsym}
\usepackage{times}
\usepackage{booktabs}
\usepackage{lipsum}
\usepackage{multirow}
\usepackage{graphicx}
\usepackage{longtable}
\usepackage{hyperref}
\usepackage{comment}
\usepackage{graphicx}
\usepackage{subcaption}
\usepackage{mwe}
\renewcommand{\thefootnote}
{\fnsymbol{footnote}}

\usepackage[T1]{fontenc}
\usepackage[utf8]{inputenc}

\usepackage{microtype}

\usepackage{inconsolata}

\title{Asking Clarification Questions to Handle Ambiguity in Open-Domain QA}

\author{Dongryeol Lee\textsuperscript{1$\ast$} \hspace{1cm} Segwang Kim\textsuperscript{2$\ast$$\ddagger$} \hspace{1cm} Minwoo Lee\textsuperscript{1} \\
{\bf Hwanhee Lee\textsuperscript{3}} \hspace{1cm} {\bf Joonsuk Park\textsuperscript{4,5,6}} \hspace{1cm} 
{\bf Sang-Woo Lee\textsuperscript{4,5,7}} \hspace{1cm}{\bf Kyomin Jung\textsuperscript{1$\dagger$}}\\
  \textsuperscript{1}Dept. of ECE, Seoul National University, \textsuperscript{2}Samsung Electronics Mobile eXperience, \\\textsuperscript{3}Chung-Ang University, 
  \textsuperscript{4}NAVER AI Lab, 
  \textsuperscript{5}NAVER Cloud, \\
  \textsuperscript{6}University of Richmond, 
  \textsuperscript{7}KAIST AI\\
  \texttt{\{drl123, ksk5693, minwoolee, kjung\}@snu.ac.kr},
  \texttt{hwanheelee@cau.ac.kr}\\
  \texttt{park@joonsuk.org},
  \texttt{sang.woo.lee@navercorp.com} \\}

\begin{document}
\maketitle

\footnotetext{\textsuperscript{$\ast$} Equal contribution.}
\footnotetext{\textsuperscript{$\dagger$} Corresponding authors.}
\footnotetext{\textsuperscript{$\ddagger$} Work done while he was in Seoul National University.}

\renewcommand*{\thefootnote}
{\arabic{footnote}}
\setcounter{footnote}{0}

\begin{abstract}
Ambiguous questions persist in open-domain question answering, because formulating a precise question with a unique answer is often challenging.
% Previously, \citet{min-etal-2020-ambigqa} have tackled this issue by generating disambiguated questions for all possible interpretations of the ambiguous question. This can be effective, but not ideal for providing an answer to the user.
Previous works have tackled this issue by generating and answering disambiguated questions for all possible interpretations of the ambiguous question.
Instead, we propose to ask a clarification question, where the user's response will help identify the interpretation that best aligns with the user's intention.
We first present \textsc{CAmbigNQ}, a dataset consisting of 5,653 ambiguous questions, each with relevant passages, possible answers, and a clarification question. The clarification questions were efficiently created by generating them using InstructGPT and manually revising them as necessary.
% To support research on this task, we present strong baseline models and evaluation metrics for \textsc{CAmbigNQ}. 
% % We also introduce new metrics for each task, which are designed to evaluate the quality and effectiveness of the generated clarification questions.
% % Through intrinsic and extrinsic evaluations of various models, we demonstrate the efficacy and challenges of the task and the dataset.
We then define a pipeline of three tasks---(1) ambiguity detection, (2) clarification question generation, and (3) clarification-based QA. In the process, we adopt or design appropriate evaluation metrics to facilitate sound research. 
% \textcolor{red}{
% Our best end-to-end performance of 27.3 F1 demonstrates the promising but difficult nature of using clarification questions to handle ambiguity. This in turn showcases the need for further efforts supported by resources like ours.\footnote{The data and code for this work will be made available at \url{TBA}.} 
Lastly, we achieve F1 of 61.3, 25.1, and 40.5 on the three tasks, demonstrating the need for further improvements while providing competitive baselines for future work.
%We achieve 40.5 F1 on clarification-based QA and 27.3 F1 on end-to-end experiments, providing strong baselines for future work.
% Notably, the preference test result shows a clear preference for CQ over DQ, aligning with our initial motivation and objectives.
% }

\end{abstract}

\section{Introduction}
\begin{figure}[t]
\centerline{\includegraphics[width=\columnwidth, trim={0cm 0cm 0cm 0cm}, clip]{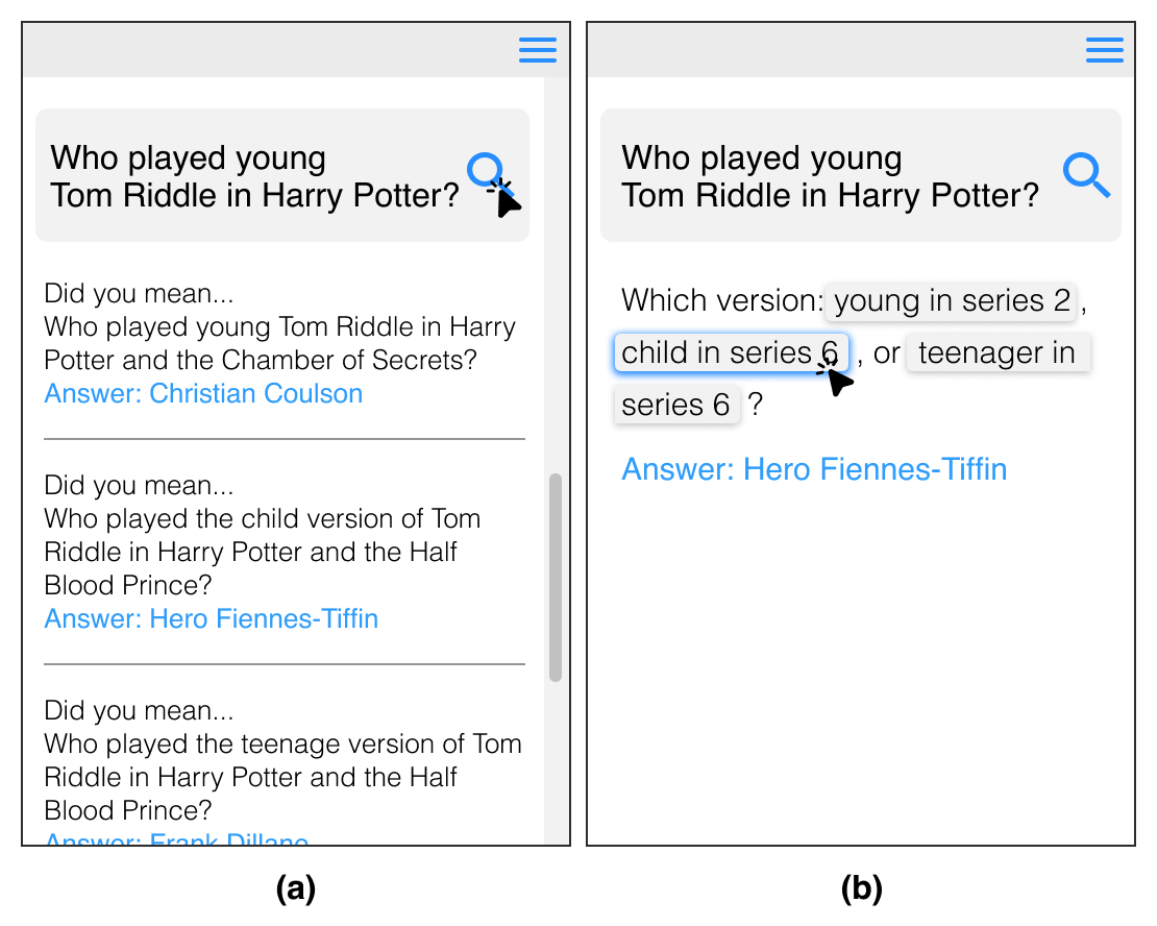}}
\caption{Two possible approaches to handling ambiguous questions (AQs) in open-domain question answering (ODQA): (a) presenting disambiguated questions (DQs) with answers (following \citet{min-etal-2020-ambigqa}), and (b) asking a clarification question (CQ) and displaying an answer based on the user's response to the CQ (ours). 
%{} \todo{Some changes we should make to the figure:
% Find an example where the category isn't "one"/
% "DQ1" -> "Did you mean ..."/
% "DQ2" -> "Did you mean ..."/
% "DQ3" -> "Did you mean ..."/
% recolor all "Answer: [answer]" to blue in the DQ case as well./
% Lastly, it may look better if we move "Disambiguated Question" and "Clarification Question" to the bottom and change them to "(a) Disambiguated Questions" and "(b) Clarification Question"
% }
%The machine reads relevant documents about the user’s question, identifies its ambiguous part, and suggests possible interpretations.
}
\label{fig:amb_ex}
\vspace*{-0.3cm}
\end{figure}

% Ambiguities in users’ questions often emerge in open-retrieval question answering (QA) \cite{min-etal-2020-ambigqa, zhang-choi-2021-situatedqa}. For example, as shown in Figure 1, a user can ask “Who won the most Grammys of all time”. 
% The question, seemingly clear and well-formulated, suddenly becomes ambiguous after the evidence documents are retrieved; “Who” could mean either “individual", “group”, or “producer”. 
%Thus, one needs to design a clever method to resolve these ambiguities.
In open-domain question answering (ODQA), questions can often be interpreted in several ways, each with a distinct answer~\cite{min-etal-2020-ambigqa, zhang-choi-2021-situatedqa}. For example, consider the question at the top of Figure~\ref{fig:amb_ex}.
% The question, seemingly clear and well-formulated, suddenly becomes ambiguous after the evidence documents are retrieved; 
Though the question seems unambiguous, ``\textit{young Tom Riddle}'' can mean “\textit{young version in series 2}'', “\textit{child version in series 6}'', or “\textit{teenager version in series 6}''.
Such ambiguity needs to be resolved to correctly find the answer sought by the user.

Previous studies propose to handle ambiguous questions (AQs) by generating a disambiguated question (DQ; disambiguated variation of the given AQ) for each possible interpretation~\cite{min-etal-2020-ambigqa,gao2021answering,stelmakh2022asqa}.
% }
While such DQ-based approaches are an important step toward resolving ambiguities in ODQA, imagine how it would be deployed in real life; without knowing the user's intention, the QA system would have to list all possible 
%DQs and respective 
answers to the user, as shown in Figure~\ref{fig:amb_ex}(a). This is not suitable in most real-world scenarios where QA systems communicate with their users through speech or small-screen devices~\cite{zamani2020mimics, croft2019importance, culpepper2018research}. 
% Also, users are unlikely to prefer all the extraneous information, as confirmed by our preference test. 

% % On the other hand, asking a single \textit{clarification question}, such as “Could you clarify ‘Who’-’which individual’, ‘which group’ or ‘which producer’?”, would be ideal.
% % Indeed, such clarification questions can provide not only functional benefits like grasping users’ intents but also emotional benefits like improving user’s satisfaction or confidence with the machine \cite{zamani2020generating}. 
% % However, few studies have been carried out for generating clarification questions in open-retrieval QA due to the lack of data and evaluation methods.
% Instead, we define the clarification question generation task for ODQA which prompts the user with a clarification question (CQ), as shown on the right side of Figure~\ref{fig:amb_ex}. More specifically, given an ambiguous question and relevant passages, the goal is to generate a clarification question consisting of the possible interpretations (\textit{options}, e.g. ``the individual") along with the general term of the entire options (\textit{category}, e.g. ``which one").
% The user's response to the question should help narrow down the set of possible answers to a single answer. 
% This approach is not only appropriate in the aforementioned real-world scenarios but also may improve the overall user experience~\cite{zamani2020generating}. 
Instead, we propose to prompt the user with a clarification question (CQ), as shown in Figure~\ref{fig:amb_ex}(b). 
% \textcolor{red}{
More specifically, given an AQ, the goal is to ask a CQ consisting of the possible interpretations as \textit{options} (e.g. ``teenager in series 6'') along with a \textit{category} summarizing the options (e.g. ``version'').
% }
Ideally, the user's response to the CQ would help identify the interpretation that best aligns with the user's intention, and the corresponding answer can be presented to the user. This CQ-based approach is not only applicable in the aforementioned real-world scenarios, but also shown to be preferred by users according to our preference test. This is also consistent with the finding that asking CQs can improve user experience with "limited bandwidth" interfaces~\cite{zamani2020generating}. 

% In this paper, we first construct \textsc{CAmbigNQ} (Clarifying Ambiguous Natural Questions), an open-retrieval QA dataset consisting of ambiguous questions and their clarification questions, and propose a clarification question generation task. 
% To secure \textsc{CAmbigNQ} at no cost, we design an auto-conversion scheme and apply it to an existing human-annotated QA dataset, \textsc{AMBIGNQ} \cite{min-etal-2020-ambigqa}. 
% To assess generated clarification questions, we propose two evaluation methods that use intrinsic references in \textsc{CAmbigNQ} and an external neural QA model, respectively. 
% Then, we establish deep learning baselines for our task and extrinsic evaluation.
% Our analysis shows that our task is challenging, and generated clarification questions can be used to revise the ambiguous questions.
 
To support research on CQ-based approaches to handle AQs in ODQA,
we present Clarifying Ambiguous Natural Questions (\textsc{CAmbigNQ}).
\textsc{CAmbigNQ} is a dataset consisting of 5,653 AQs from \textsc{AmbigNQ}~\cite{min-etal-2020-ambigqa}, each accompanied by relevant passages, possible answers, and a newly created CQ. 
The CQs were first generated using InstructGPT~\cite{Ouyang2022TrainingLM} through in-context few-shot learning, then manually vetted and edited as necessary by human editors. Such human-machine collaboration for constructing corpora has been shown to significantly reduce the time and the cost from fully manual approaches~\cite{Wang2021WantTR, Wu2021ASO}.

% We propose the method for building a dataset that leverages the in-context few-shot learning ability of Large-scale Language Models (LLMs) \cite{Brown2020LanguageMA}, followed by the human validation process.
% Our method allows for a significant reduction in the cost of constructing the entire dataset through manual annotations. 
% \textsc{CAmbigNQ} was constructed from the set of ambiguous questions, and respective disambiguated questions and answers, in an existing dataset, \textsc{AmbigNQ} \cite{min-etal-2020-ambigqa}.

% \textcolor{red}{
% We also define a pipeline of three tasks to handle AQs in ODQA: (1) ambiguity detection, (2) clarification question generation, and (3) clarification-based QA.
% Furthermore, we develop task-specific evaluation metrics to assess the performance and effectiveness of each task within the pipeline.
% In the experiments, we achieve 40.5 F1 on clarification-based QA (27.3 F1 end-to-end), forming strong baselines for future work.
% }
We also define a pipeline of three tasks to handle AQs in ODQA---(1) ambiguity detection, (2) clarification question generation, and (3) clarification-based QA. In the process, we  adopt or design appropriate evaluation metrics to facilitate sound research. 
% \textcolor{red}{
The experiments show that though they were shown to be helpful for generating DQs, predicted answers for AQ do not help improve the CQ-based approach overall.
Lastly, we achieve F1 of 61.3, 25.1, and 40.5 on the three tasks, demonstrating the need for further improvements while providing competitive baselines for future work.\footnote{The data and code will be available at \url{https://github.com/DongryeolLee96/AskCQ}}

Our main contributions are threefold:
\begin{itemize}
\vspace{-1.5mm}
\item We propose to use CQs as a practical means to handle AQs in ODQA. Consistent with the findings by ~\citet{zamani2020generating}, our human preference test shows that \textit{the use of CQ is preferred over that of DQs
}(Section \ref{sec:preference}).
% (RQ1;\S\ref{sec:preference})}.
\vspace{-1mm}
\item We present \textsc{CAmbigNQ}, a dataset to support CQ-based handling of AQs in ODQA. It was built efficiently by leveraging a well-curated resource, \textsc{AmbigNQ}, as well as the power of InstructGPT and human editors
(Section \ref{sec:dataset}). 
\vspace{-1mm}
% a dataset constructed by combining the power of InstructGPT and human editors to generate clarification questions (CQs) for ODQA. 
% a dataset consisting of 5,654 ambiguous questions, each accompanied by relevant passages, possible answers (from \textsc{AmbigNQ}), and a newly created clarification question.

\item We define a pipeline of tasks and appropriate evaluation metrics for handling AQs in ODQA
(Section \ref{sec:task}). 
% and provide competitive baselines. 
\textcolor{black}{
The experiments show that though they were shown to be helpful for generating DQs, \textit{predicted answers for AQ do not help improve the CQ-based approach overall} (Section \ref{sec:experiments}).
% (RQ2; \S\ref{sec:experiments})}.
% The experiments show that the utilization of only question and relevant passages in the CQ pipeline leads to the highest performance.
}
% \vspace{-3mm}
% \item We introduce strong baseline models for \textsc{CAmbigNQ} with evaluation metrics that are specifically designed to provide a more comprehensive and accurate assessment of the quality of the generated CQs.\todo{check}

\end{itemize}
\section{Related Work}

% In this section, we provide an overview of existing work on CQ generation in various domains, along with the methods of constructing datasets and their ways of evaluating the outputs. 

\paragraph{Clarification Question Datasets}
% To resolve question ambiguity, CQs have been introduced and related datasets have been released in various domains. 
% Our dataset was built upon an \textsc{AmbigNQ} \cite{min-etal-2020-ambigqa}, the dataset consists of AQs, each accompanied by DQs, and corresponding answer sets for DQs.
To resolve question ambiguity, CQ datasets have been released in various domains. 
% In information seeking setting, \citet{zamani2020mimics} propose MIMICS, a collection of search clarification dataset for real web search domain and \citet{aliannejadi2019asking} introduce Qulac dataset, which is the first clarifying questions dataset in conversational domain.
In the information-seeking domain, CQ datasets for conversation \cite{aliannejadi2019asking, aliannejadi2020convai3, Guo2021AbgCoQACA, wu2022inscit} or
web search \cite{zamani2020mimics} have been collected from crowdsourcing or real users' follow-up search queries.
In the question-answering domain, datasets that focus on specific topics \cite{rao-daume-iii-2018-learning, braslavski2017you, saeidi-etal-2018-interpretation} or knowledge-base \cite{xu-etal-2019-asking} has been proposed. 
To the best of our knowledge, we are the first to release a CQ dataset for ODQA.\footnote{\citet{xu-etal-2019-asking} presents a CQ dataset for KBQA that is also open-domain, but the setting is much more restricted than ODQA in that the goal is to find the appropriate knowledge-base entry between exactly two entries about the same entity. Once an entry is determined, answering the question involves a simple table lookup, rather than a full-on QA.}

% In knowledge-based QA domain, \citet{xu-etal-2019-asking} construct CLAQUA, which is CQ dataset that are intentionally annotated with pre-specified entity reference ambiguities. 
%  \citet{zamani2020mimics} proposed MIMICS, a collection of search clarification dataset for real web search queries proposed a template-based and a sequence-to-sequence generative model to produce clarifying questions \citep{zamani2020generating}.
% \citet{aliannejadi2019asking} introduced Qulac dataset, which is the first clarifying questions dataset in an open-domain information-seeking conversational setting and later released ClariQ which consists of the data that is more suitable for conversational setting\citep{Aliannejadi2020ConvAI3GC}. 
% Research on community QA domain\citet{rao-daume-iii-2018-learning, Rao2019AnswerbasedAT, braslavski2017you} focuses on finding underspecified questions in the network of online question answering websites.
% In knowledge-based QA domain, \citet{xu-etal-2019-asking} construct CLAQUA, which is CQ dataset that are intentionally annotated with pre-specified entity reference ambiguities. 
% To the best of our knowledge, there is no existing clarifying question dataset in open domain settings. 

\paragraph{Dataset Construction Leveraging LLMs}
%Previous studies on constructing CQ datasets mostly depend on crowdsourcing or real users' search queries.
% When we expand this scope into the question-answering domain, still most datasets depend on crowdsourcing \cite{Yang2018HotpotQAAD, Rajpurkar2016SQuAD1Q} or real users' search queries \cite{Campos2016MSMA, Kwiatkowski2019NaturalQA}.
Manually constructing datasets from scratch is laborious and costly, which can be prohibitive depending on the nature of the dataset. Also, access to real users' data is strictly limited to a certain community.
To mitigate these issues, approaches leveraging LLMs to construct datasets have recently been used in various domains such as dialogue~\cite{Bae2022BuildingAR}, domain-adaptation~\cite{Dai2022PromptagatorFD}, and in general~\cite{Ding2022IsGA}.
However, such an approach has not been used to construct CQ datasets, except for ours.
%Nevertheless, we are the first to propose using this approach followed by human validation on CQ data generation for the first time.
We used InstructGPT ~\cite{Ouyang2022TrainingLM} to generate CQs through in-context few-shot learning, and then manually vetted and edited them as necessary to construct our dataset.

\begin{figure*}[t]
\centerline{\includegraphics[width=0.9\textwidth]{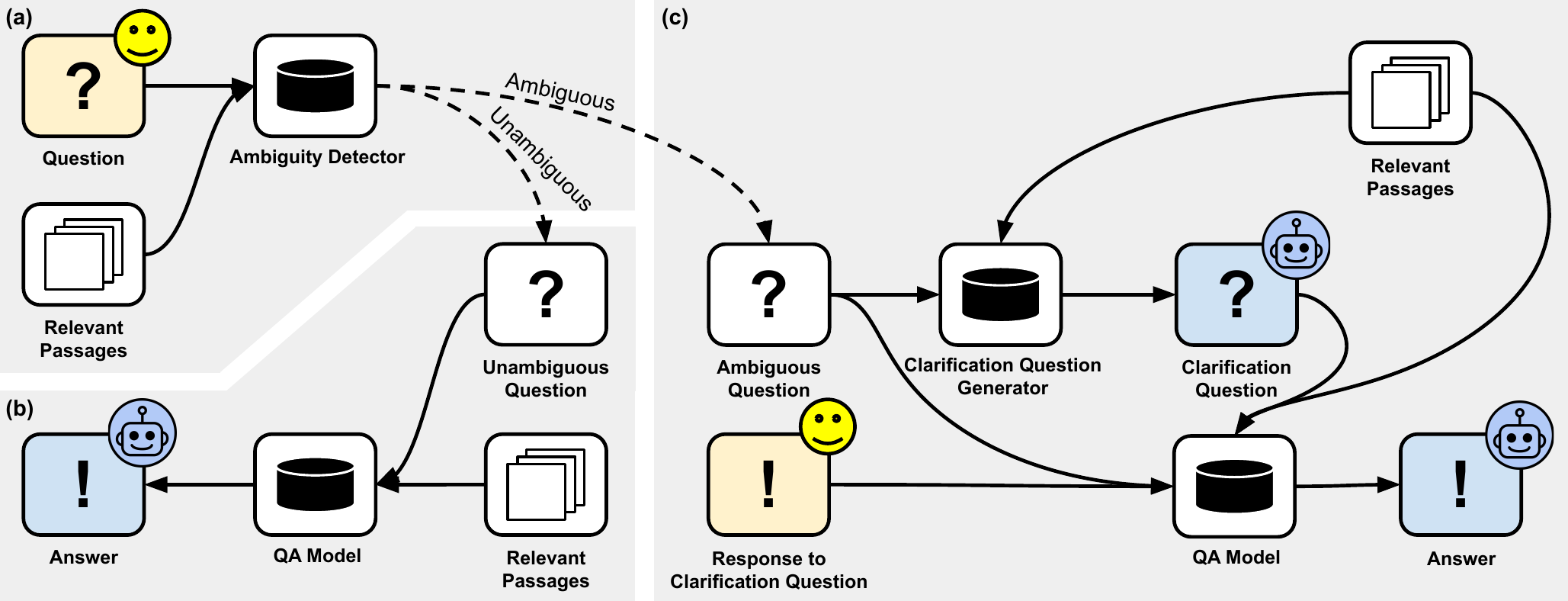}}
\caption{
Overview of our proposed approach to ODQA. Given a question, it is first checked for ambiguity ((a) ambiguity detection). 
If it is not ambiguous, it is processed in a normal QA setup ((b) QA; outside the scope of this work).
Otherwise, an extra step of eliciting a response to a clarification question precedes QA ((c) CQ generation + clarification-based QA).
 Yellow blocks represent the user input, and  blue blocks, the system output.
}
\label{fig:overview}
\vspace*{-0.3cm}
\end{figure*}
 
\paragraph{Clarification Question Evaluation}
%CQ 의 quality 를 평가하는 것은 왜 어렵냐? -> 정해진 답이 없기 때문에? 예를 들어서? 
% automatic measurement metric 들을 나열하고 자마니가$
There are several options for evaluating the quality of CQs, 
First is leveraging widely-used automatic text evaluation metrics, such as BLEU \cite{Papineni2002BleuAM} or ROUGE \cite{Lin2004ROUGEAP}.
However, due to the poor correlations between such scores and human evaluation, \citet{zamani2020mimics} strongly discourages the use of such metrics for evaluation.
% since references are not the only ground-truth and user satisfaction poorly correlates with resulting scores.
Second is human evaluation. While it typically provides a reliable estimate of how people would think of the given CQs, it can be time-consuming and costly.
As a third option, evaluation methods using external neural models have recently been introduced~\cite{Rei2020COMETAN, Mehri2020USRAU, Lee2021KPQAAM}. This approach improves on the first, without the burden of the second. 
In this work, we design evaluation methods that suit our tasks, leveraging external neural models to provide a more comprehensive and accurate assessment.

\section{Task Overview}
\label{sec:task}
We propose to handle AQs in ODQA by asking CQs as shown in Figure~\ref{fig:overview}. There are three tasks: (1) \textit{ambiguity detection}, (2) \textit{clarification question generation}, and (3) \textit{clarification-based QA}.

% \textcolor{red}{Subsequent to defining the tasks, we conduct a human preference test to evaluate the efficacy of our approach in comparison to previous disambiguation methods. 
% }
%This evaluation serves to demonstrate the superiority of our proposed approach in enhancing user experience and effectively handling ambiguous questions within the ODQA framework.
% In the remainder of this paper, we adopt the following notations: the reference CQ $q_c$ contains the category $cat$ and the options $O = \{o_1, \dots, o_N\}$, and the predicted CQ $\hat{q_c}$ contains the category $\hat{cat}$ and the options $\hat{O} =\{\hat{o}_1, \dots, \hat{o}_M\}$, respectively, where $N$ and $M$ may differ.

\subsection{Task 1: Ambiguity Detection}
\paragraph{Task}
% The ambiguity detection task is defined as follows:
Given a question and relevant passages, the goal is to determine whether the question is ambiguous or not, as shown in Figure~\ref{fig:overview}(a).
% The goal of ambiguity detection is to determine whether a given question is ambiguous or not. 
A question is considered ambiguous if it can be interpreted in multiple ways, each of which is associated with a unique answer, resulting in multiple possible answers. 
A question is unambiguous if it can only be interpreted in one way, thus one possible answer. 
% \textcolor{red}{(e.g. Who is the president of United States as of 2022?)}
% More formally: This is a binary classification task with a question and relevant passages as input, and \textit{Ambiguous} or \textit{Unambiguous} as two possible outputs.

%Questions with a unique interpretation are defined as Single-interpreations Questions (SQs), while those with multiple interpretations are defined as AQs.
%Since there is no need for CQs for the questions with a single answer, the questions with single answers do not proceed to the next step in the pipeline.

\paragraph{Evaluation}
% Our \textit{Ambiguity Detection Evaluation} tests whether the model can correctly identify questions that have a single answer versus those that have multiple answers. 
For this binary classification task, we use standard metrics, such as accuracy, recall, precision, and F1.
% We measure the recall and precision of the model's output using standard definitions of these metrics.
% In this context, recall is a measure of the model's ability to correctly identify all of the questions with multiple answers, while precision is a measure of the model's ability to correctly identify only the questions with multiple answers, without mistakenly labeling any questions with a single answer as having multiple answers.

\subsection{Task 2: Clarification Question Generation}
\label{sec:cqg}
\paragraph{Task}
% The CQ generation task is defined as follows:
Given an AQ and relevant passages, the goal is to generate a CQ such that a valid response to the CQ is associated with exactly one of the multiple answers to AQ.

A CQ is typically formatted as follows:

% ``\textit{Which $[category]$: $[option_1], ...,$ or $[option_n]$?}''
``\textit{Which $[category]$: $[option_1], [option_2], ...,$ or $[option_n]$?}''

Here, $[category]$ is a category to which all options belong, such as ``version'' in 
Figure~\ref{fig:amb_ex}.  
If the options can not be grouped into a single category, ``one'' is used as a placeholder for the category. 
Also, where suitable, additional words like prepositions can precede ``Which,'' e.g. ``In which context''.
There should be an $[option_j]$ for each possible interpretation of the AQ. Also, only the options are considered valid responses to the given CQ.

\paragraph{Evaluation} 
We evaluate the quality of the generated CQ in two levels of granularity. 
First, we compare the generated CQ against the reference CQ using the standard BLEU-4 metric and \textsc{Bertscore}~\cite{zhang2019bertscore}.

% \textcolor{red}{
Second, we evaluate the category and the options separately for a more fine-grained evaluation. 
For the category, exact match (EM) and BLEU-1 are computed since the category is typically very short.
% }
% We also perform a fine-grained evaluation to evaluate the quality of the category part and options part of the generated CQ separately.
For the options, we adopt and adjust the partial match evaluation metric by \citet{Li2022MultiSpanQAAD}, whose goal is to measure the similarity between a predicted set and a reference set of strings. 
Since the exact alignment of the strings between the sets is unknown, it measures the similarity---based on the longest common substring (LCS)---between all pairs across the sets and keeps the highest score. Here, multiple strings from the predicted set can be aligned with the same string in the reference set.
% \textcolor{red}{
In this work, we impose a constraint that limits the alignment of a reference option to at most one predicted option, since each option should represent a unique interpretation of the AQ.
% }
% In our work, we introduce the following constraint: a reference option can be aligned to at most one predicted option. This is because each option should represent a unique interpretation of the AQ. 
Thus, we find the optimal assignment that maximizes the similarity score using the Hungarian algorithm \cite{Kuhn1955TheHM} and compute precision, recall, and F1 as follows:
\vspace{-3pt}
\begin{equation}
\begin{aligned}
\textstyle{\text{max}_{i}^{prec} = \sum_{p\in P_i} sim(p, f_{i}(p))},
\end{aligned}
\label{eq:prec}
\end{equation}
\vspace{-25pt}

\begin{equation}
\begin{aligned}
\textstyle{\text{max}_{i}^{rec} = \sum_{r\in R_i} sim(r, f_{i}^{-1}(r))},
\end{aligned}
\label{eq:rec}
\end{equation}
\vspace{-25pt}

\begin{equation}
\begin{aligned}
\text{prec}=\frac{ \sum_{i} \text{max}_{i}^{prec}}{\sum_{i} |P_i| },
\text{rec}=\frac{ \sum_{i} \text{max}_{i}^{rec}}{\sum_{i} |R_i| },
\end{aligned}
\label{eq:prec_and_rec}
\end{equation}
\vspace{-2pt}
% \begin{equation}
% \begin{aligned}
% \text{F1}=\frac{2 \times \text{prec} \times \text{rec}}{\text{prec}+\text{rec}},
% %\textstyle{\text{F1}=\frac{(2 \cdot \text{Precision} \cdot \text{Recall})}% {({\text{Precision} + \text{Recall})}}}
% \end{aligned}
% \label{eq:f1}
% \end{equation}
where $P_{i}$ and $R_{i}$ is the set of predicted and reference options for $i$-th sample, $sim(\cdot)$ is the LCS-based similarity measure, and $f_{i}:P_{i} \rightarrow R_{i}$ is the optimal one-to-one mapping computed from the Hungarian algorithm. F1 is a harmonic mean of precision and recall, as usual.

Please refer to Appendix~\ref{app:intrinsic} for more details.
%on the similarity metric $sim$ and optimal one-to-one mapping $f_{i}$.
%on calculating $\text{max}^{prec}$ and $\text{max}^{rec}$.

\subsection{Task 3: Clarification-based QA}
\label{sec:cbqa}

\paragraph{Task}
% The Clarification-Based Question Answering task for ODQA is defined as follows:
% Given an AQ, relevant passages, and the predicted CQ, the goal is to generate a unique answer for each corresponding interpretation of the AQ.

Given an AQ, relevant passages, and a CQ, the goal is to generate a unique answer for every valid answer to the CQ---i.e., an option---which is associated with an interpretation of the AQ.%---thus a unique answer to---the AQ.

Each answer is generated by calling a QA model on an \textit{AQ revised by CQ}, which is the concatenation of AQ, category, and single option: 
% $AQ, \hat{cat}: \hat{o}_m$. 

``$AQ$\textit{, Which $[category]$: $[option_j]$}''.

Note, because each \textit{AQ revised by CQ} is a unique interpretation of the AQ with a distinct answer, the relevant passages first need to be reranked before generating an answer.\footnote{We utilized cross encoder MiniLMv2 model~\cite{wang2020minilmv2} fine-tuned on MSMARCO. \url{https://huggingface.co/cross-encoder/ms-marco-MiniLM-L-12-v2}}

% To do this, the CQ is first split into individual interpretations by revising AQ into a specific form.
% The format of AQ revised by CQ is as follows: ``\textit{AQ, [category]: [option\_i]?}", where \textit{[option\_i]} is a single option from the CQ that leads to corresponding \textit{[interpretation\_i]}.
% Because the relevant passages were retrieved based on the AQ, which has different interpretations from each specific option, they are reranked based on each AQ revised by CQ. 
% The reader model is then provided with the AQ revised by CQ and the reranked relevant passages as input, and its ability to output the corresponding answer is evaluated.

\paragraph{Evaluation}
% The predicted CQs are tested whether they can be used to disambiguate respective AQs so that a neural QA model, i.e., the reader, can output the correct answers.
% % Our extrinsic evaluation method validates the usefulness of predicted CQs using a external neural QA model, i.e., reader. 
% % The method tests whether the CQs can revise AQs so that the reader outputs unique correct answers.
% % To test the usefulness of predicted CQs for revision, we propose an extrinsic evaluation method based on a neural QA model, i.e., reader. 
% We first define AQ revised by CQ as the following form: the concatenation of AQ, category, and single option ($AQ, \hat{cat}: \hat{o}_m$).
% Since each option leads to a semantically different option, each option should have different answers from the AQ, and the corresponding relevant passages should be reordered.
% In that sense, to evaluate a predicted category $\hat{cat}$ and an option $\hat{o}_m$, we first rerank the relevant passages of AQ revised by CQ. 
% Then the reader takes the combination of AQ revised by CQ and reranked passages as input, and outputs the answer for each option $\hat{o}_m$.

% The clarification-based question answering (CBQA) evaluation 
The procedure is similar to that of option evaluation for CQ generation, in that it uses the partial match method with the Hungarian algorithm to determine the optimal alignment between predicted and reference answers.
% Since the correspondence of the answers in the predicted set and reference set is unknown, the procedure is similar to that of CQ generation option evaluation.
% We correspondingly use the partial match method with the Hungarian algorithm to determine the optimal alignment between the set of predicted answers and that of the reference answers.
% In the same manner as the evaluation of CQ generation task, we apply a specific similarity score function for the CBQA evaluation to calculate the maximum sum of the precision score ($\text{max}^{prec}$) and the recall score ($\text{max}^{rec}$) for each pair of predicted and reference answer, with a one-to-one alignment.

The only difference is that $\text{max}_{i}^{prec}$ and $\text{max}_{i}^{rec}$ for each aligned pair of predicted and reference answers are computed differently, because in QA, the correct answer may be expressed in multiple ways, e.g., ``Michael Jordan", ``MJ",  and ``Jordan". Thus, a predicted answer is compared with all variations of the same answer, and the max score is used for that pair. 
% The precision, recall, and F1 are then calculated using the maximum sum of the precision and recall scores in the same manner as Equation~\ref{eq:prec}, Equation~\ref{eq:rec}, and Equation~\ref{eq:f1}.
Then, precision, recall, and F1 are calculated as before, with the newly computed $\text{max}_{i}^{prec}$ and $\text{max}_{i}^{rec}$.

% The Hungarian algorithm is applied to each of the precision-based and recall-based overlap score matrices and
% $max^{prec}$ and $max^{rec}$ are calculated from Equation~\ref{eq:max_prec}, and Equation~\ref{eq:max_rec} respectively.
% The precision and recall for extrinsic evaluation are then calculated from Equation~\ref{eq:prec}, Equation~\ref{eq:rec} using $max^{prec}$ and $max^{rec}$, and the F1 score is measured by Equation~\ref{eq:f1}.

% Please refer to Appendix~\ref{app:extrinsic} for details on calculating the maximum sum of the precision and recall score.
Please refer to Appendix~\ref{app:extrinsic} for more details.

% \paragraph{CQ Answering}
% The CQ answering task for ODQA is defined as follows:
% Given an AQ, relevant passages, and CQ, the goal is to generate a unique answer for each corresponding interpretation of the AQ.
% To do this, the CQ is first split into individual interpretations by revising AQ into a specific form.
% The format of AQ revised by CQ is as follows: ``\textit{AQ, [category]: [option\_i]?}", where \textit{[option\_i]} is a single option from the CQ that leads to corresponding \textit{[interpretation\_i]}.
% For example, given the AQ and CQ from figure ~\ref{fig:amb_ex}, the AQ revised by CQ for the option ''the individual" would be ''Who won the most Grammys of all time, which one: the individual?". 
% Because the relevant passages were retrieved based on the AQ, which has different interpretations from each specific option, they are reranked based on each AQ revised by CQ. 
% The reader model is then provided with the AQ revised by CQ and the reranked relevant passages as input, and its ability to output the corresponding answer ''Georg Solti" is evaluated.
% \input{TABLES/cq_examples.tex}

\section{The \textsc{CAmbigNQ} Dataset}
\label{sec:dataset}

\subsection{Dataset Construction}
We present Clarifying Ambiguous Natural Questions (\textsc{CAmbigNQ}), a new dataset consisting of 5,653 AQ, each with relevant passages, possible answers, and a CQ.
\textsc{CAmbigNQ} was constructed from the AQs in \textsc{AmbigNQ}~\cite{min-etal-2020-ambigqa}, which provides each AQ with relevant passages, as well as the DQ and answer pairs reflecting the possible interpretations and respective answers of the AQ. 
To build \textsc{CAmbigNQ}, we replaced each set of DQs with a CQ. In other words, the CQ is an integrated version of the set of DQs. 
%This cause the main challenge of the annotation process that each option of CQ should represent the corresponding DQ. However, 
Representing each DQ as a single phrase option can be cumbersome to do manually.
Thus, we collect high-quality of CQs by leveraging InstructGPT, using a two-step framework: \textit{Generation via InstructGPT} and \textit{Manual Inspection and Revision}.
% \subsection{Auto-conversion Scheme}\label{subsec:auto-conversion}
% We now describe our auto-conversion scheme that outputs CQs from both AQs and DQs in \textsc{AmbigNQ}.
% Specifically, by using the off-the-shelf English model \textit{spaCy}, we first extract all \textit{[option\_i]} corresponding to each DQ of an AQ, then decide \textit{[ambiguous part]} in the AQ.
% We defer the detailed explanation of the scheme to Appendix~A. 

% We extract an \textit{ambiguous part} from an AQ based on a modification-scoring.
% We choose the part only from candidates that are the AQ's spans, such as noun chunks, VERB, and ADJ, suggested by the parser.
% To select the part from candidates, we devise the modification-scoring:
% going through each DQ, a candidate span is scored if the \textit{spaCy} tagger recognizes that \textit{option\_i} modifies the span.

% \subsection{Dataset Construction}
% \paragraph{Clarification Question Format}
% The format of the CQs is restricted to the following: ``\textit{Which [category]: [option\_1], [option\_2], ... , or [option\_n]?}"
% In this format, the \textit{category} is a general term that groups the following options, and \textit{option\_i} is a representation of each interpretation.
% In cases where the options can not be grouped into a single category, the general term ``Which one'' is used for the category. 
\paragraph{Generation via InstructGPT}
To take advantage of the few-shot learning capability of InstructGPT, we first manually annotate a small number of CQs 
%to ensure that each option represents a specific interpretation. 
for AQs.
These edited CQs are then used as "few-shot" examples along with brief instructions and both the AQ and corresponding DQs. 
We sampled \textit{six} examples considering the diversity of \textit{category} and \textit{number of options}.  
The final prompt is in the form of a concatenation of the six examples, instructions, target AQ, and target DQs in the following form:

``\textit{instruction, AQ\_1, DQs\_1, CQ\_1, ..., instruction, AQ\_6, DQs\_6, CQ\_6, instruction, AQ\_target, DQs\_target}''

% \todo{example? This section too short compared to Manual Revision}

\begin{table}[]
\centering
\small
%\resizebox{\columnwidth}{!}{%
\begin{tabular}{lccccc}
\hline
\multirow{2}{*}{Split} & \multicolumn{2}{c}{CQ} & Category & \multicolumn{2}{c}{Options}\\ \cmidrule(lr){2-3} \cmidrule(lr){4-4} \cmidrule(lr){5-6}

& \# & Len. & Len. & Avg. \# & Len.  \\ \hline
Train   & 4,699  & 13.6    & 2.8   & 2.9   & 3.3        \\
Validation & 461   & 15.9    & 2.5    & 3.3   & 3.8       \\
Test       & 493   & 17.8    & 2.8    & 3.4   & 4.1       \\ \hline
\end{tabular}%
%}

% \begin{table}[]
% \centering
% \resizebox{\columnwidth}{!}{%
% \begin{tabular}{llllll}
% \toprule
%           & \#CQs & CQ len. & AP len. & \#Ops & Op len.\\
% \hline
% % training   & 4,739	& 2.9   & 18.5	& 2.3	& 4.5\\
% % validation & 1,002	& 3.5	& 22.7	& 2.3	& 4.9\\
% training & 4739	& 18.5	& 2.3	& 2.9	& 4.5	\\
% validation & 496	& 22.8	& 2.3	& 3.5	& 4.9	\\
% test & 506	& 22.6	& 2.4	& 3.5	& 5.0	\\
% \bottomrule
% \end{tabular}
% }

\caption{
Statistics of \textsc{CAmbigNQ}.
Each clarification question (CQ) consists of one category and multiple options. 
The length is reported in the number of words.
% Avg. \# is the average number of options per clarification question.
}
\label{tab:data_statistics}
\vspace*{-0.3cm}
\end{table}
% , resulting in 4,739 (1002) training (validation) examples.
\begin{table*}[]
\renewcommand{\arraystretch}{1.1}
\centering
\small
%\resizebox{\textwidth}{!}{%
\begin{tabular}{c|l|l}
\hline
Target & Question & Example\\ 
\hline
\multirow{3}{*}{\begin{tabular}[c]{@{}c@{}}Category\\  Only \\   (12.9\%)\end{tabular}} &
  AQ &
  Who is Catch Me If You Can based on?
\\ \cline{2-3} 
 &
  GPT CQ &
  Which \textcolor{red}{one}: the 2002 film, the book, or the musical? \\ 
 &
  Edited CQ &
  Which \textcolor{blue}{version}: the 2002 film, the book, or the musical? \\ [-1pt]
  \hline 
\multirow{3}{*}{\begin{tabular}[c]{@{}c@{}}Options\\  Only\\    (19.7\%)\end{tabular}} &
  AQ &
  When did the £20 note come out? \\ \cline{2-3}  
 &
  GPT CQ &
  Which series: F, or E? \\
 &
  Edited CQ &
  Which series: F, \textcolor{blue}{E variant,} or E? \\ [-1pt]
  \hline
\multirow{3}{*}{\begin{tabular}[c]{@{}c@{}}Category\\ \& Options\\      (31.4\%)\end{tabular}} &
  AQ &
  Who plays Will on The Bold and Beautiful? \\ \cline{2-3} 
 &
  GPT CQ &
  Which \textcolor{red}{time period}: \textcolor{red}{first}, \textcolor{red}{replacement}, or \textcolor{red}{2013}? \\
 &
  Edited CQ &
  Which \textcolor{blue}{one}: \textcolor{blue}{first actor}, \textcolor{blue}{actor that replaces the wardens},   or \textcolor{blue}{actor that began playing in} 2013? \\ [-1pt]
\hline
\multirow{3}{*}{\begin{tabular}[c]{@{}c@{}}Whole\\ Question\\      (7.8\%)\end{tabular}} &
  AQ &
  Who is the all-time passing leader in the NFL? \\ \cline{2-3} 
 &
  GPT CQ &
  \textcolor{red}{Does the leader include regular season stats, or stats from the playoffs as   well?} \\
 &
  Edited CQ &
   \textcolor{blue}{In which context: in the regular seasons, or including the playoffs as well?} \\ [-1pt]
  \hline
\multirow{3}{*}{\begin{tabular}[c]{@{}c@{}}None\\ (26.7\%)\end{tabular}} &
  AQ &
  Who is the current chairman of African Union commission?
 \\ \cline{2-3} 
 &
  GPT CQ &
   Which chairman: 4th, 3rd, or 2nd? \\
  &
  Edited CQ & Which chairman: 4th, 3rd, or 2nd?
  \\ [-1pt]
  \hline
\end{tabular}%
%}
\caption{
Examples of manual revisions made to clarification questions (CQs) generated by InstructGPT for ambiguous questions (AQs). The human editors were provided with disambiguated questions (DQs) for reference. 
% Revised tokens are colored in red and added tokens after revision are colored in blue.
Red and blue words represent the words before and after revision, respectively.
The remaining 1.5\% was marked as ``unambiguous'' by the editors, meaning only one interpretation, and thus one answer, exists for the given question. These were excluded from our dataset.
% \todo{ For ``Category'', it will be better if we can find an example where the actual category(e.g. ``year'') is edited. Also, I wonder if it makes sense to include ``no revision" case. (It wouldn't look good if the percentage is too big, though)}
}
\label{tab:changes_by_human}
\vspace*{-0.3cm}
\end{table*}

\paragraph{Manual Inspection and Revision}
% \paragraph{Qualification stage}

% Before the manual editing stage, we first recruit undergraduate students who have English-related majors and subject them to a qualification stage in order to filter out applicants.
% This stage includes instructions, 25 annotated examples done by a co-author, and 25 test cases that each applicant had to complete. 
% Out of the total number of applicants, 7 passed the qualification stage and moved on to the manual editing stages, while 3 failed to pass.

% \paragraph{Manual Editing}
The recruited annotators were asked to read instructions and revise 25 CQs accordingly as a qualification test.
Then the editors who passed the qualification test were asked to examine, and revise as necessary, the CQs generated by InstructGPT. 

% For each CQ, they were also provided with the respective AQ and DQ-answer pairs.
% In the manual editing stage of the data construction process, editors are tasked with reviewing the CQs generated by the InstructGPT.
% For each page, they are provided with the AQ, DQs, and the corresponding answer sets, along with the CQ generated by the InstructGPT. 
They were asked to follow the following protocol to ensure the quality of the final CQs: First, check whether the AQ had at least two distinct interpretations and corresponding DQs; Second, check whether the CQ generated by InstructGPT is in the correct format (See Section~\ref{sec:cqg});
Third, check whether each option accurately represents its corresponding DQ and the category is a correct term describing the set of options. 
The editors had three actions to choose from---they could either: mark the AQ as not ambiguous, i.e. there is only a single interpretation and answer (occurred in 1.5\% of cases), revise the CQ (occurred in 71.8\% of cases), or leave the CQ as is (occurred in 26.7\% of cases). (See Table~\ref{tab:changes_by_human} for example revisions.)
The high revision rate suggests that the few-shot generation with InstructGPT is far from perfect, and manual editing was necessary.

% \textcolor{red}{
For inter-annotator agreement, we use the validation set which was annotated by two annotators, following \citet{min-etal-2020-ambigqa}.
The kappa coefficient~\cite{cohen1960coefficient} is 0.623, which can be considered a "substantial agreement.''~\cite{mchugh2012interrater}.
% }
%We use the validation set, which was annotated twice by the editors, to measure the inter-annotator agreement. 

%, which involves summing scores for each class and then averaging these sums across all classes.
 
%\todo{Incomplete}

% cq_bleu = 65.8
% cat_bleu = 63.8
% option_blue = 66.0

% cq_em = 39.3
% cat_em= 56.8
% option_em= 63.9

% \paragraph{}
% Please refer to Table~\ref{tab:data_statistics} for the statistical summary of \textsc{CAmbigNQ}.

\subsection{Dataset Analysis}
% \paragraph{Overall Statistics.}
The entire dataset consists of 5,653 data points, as shown in Table~\ref{tab:data_statistics}.
%which were randomly split into train, validation, and test sets, as shown in Table~\ref{tab:data_statistics}.
% The entire dataset consists of 5,654 data points, which were obtained from the publicly available \textsc{AmbigNQ} dataset.
The training set was sourced from that of \textsc{AmbigNQ}, while the validation and test sets were randomly split from the development set of \textsc{AmbigNQ}. 
%The distribution of each set is shown in Table~\ref{tab:data_statistics}.
Each AQ in the dataset has over three interpretations on average, which in turn means that each CQ has over three options on average. 
The average length of the CQs varies from one  split to another, with a general trend of longer CQs having more options. 
% This indicates that the average length of the CQs is significantly influenced by the number of options they contain.

% \paragraph{Categories distribution.}
% In the CQ, the category and the options are separated by a colon. 
% We use a simple rule-based scheme to split the CQs into categories and options, and then compiled statistics on the categories present in each split of the dataset.

%Also, Table~\ref{tab:category_statistic} shows that the three most frequently occurring categories are consistent across all three sets, with ``season'' being the fourth most frequent category in both the test and the training set, and fifth in the validation set.
%without introducing a noticeable skew in the distribution in certain splits.
% This suggests that data points with multiple interpretations are concentrated in certain categories. 
% The high usage rate of the category "Which one" in all of the data suggests that there is a complex distribution of options within this category, making it difficult to generate appropriate options for each AQ.

% \paragraph{Common Revisions by Human Editors.}
The first column in Table~\ref{tab:changes_by_human} shows the statistics on which components of the CQs generated by InstructGPT were revised by human editors.
Of the entire dataset, about 8\% were due to the invalid format of the CQs.
This means that although InstructGPT was provided with six example CQs in the prompt, it is not always enough to enforce the format.
Additionally, one common type of revision made to the category was converted to or from ``one'', meaning InstructGPT often tried to group ungroupable options or chose not to group options that can be grouped into a single category.
A common revision made to the options was to split what InstructGPT generated as a single option. Errors like this also lead to a mismatch between the number of DQs, or interpretations, and that of the options. 
% We suspect that explicitly providing the number of interpretations in the prompt may have helped in this regard. 
Overall, there seems to be room for further prompt engineering to minimize errors, but we believe manual revisions are a necessary component for constructing high-quality datasets as of yet.
% \textcolor{red}{Please refer to Appendix~\ref{app:data_collection} for more details of the data collection process and dataset.}
Please refer to Appendix~\ref{app:data_collection} for more details.

\section{Experiment 1: CQ vs DQ Preference}
\label{sec:preference}

We first conduct a human preference test to investigate the question: \textit{Is our CQ-based approach preferred over a DQ-based one to handle AQs in ODQA?} This is to check if it is worthwhile to pursue the CQ-based approach. 
% our proposed approach is indeed preferred, as suggested by~\citet{zamani2020generating}.

% Prior to the main experiments to design \textcolor{red}{most effective} model for answering AQs using CQs, we conduct a human preference test to confirm that our proposed approach is indeed preferred, as suggested by~\citet{zamani2020generating}. 

% \textcolor{red}{Before commencing the experimental phase aimed at evaluating the initial performance of CQ, a preliminary human preference test is conducted to assess the effectiveness of both CQ and DQ. The main objective of this test is to acquire empirical evidence that highlights the effectiveness of CQ compared to DQ, particularly in terms of enhancing the overall user experience.}

\begin{figure}[t]
\centerline{\includegraphics[width=\columnwidth]{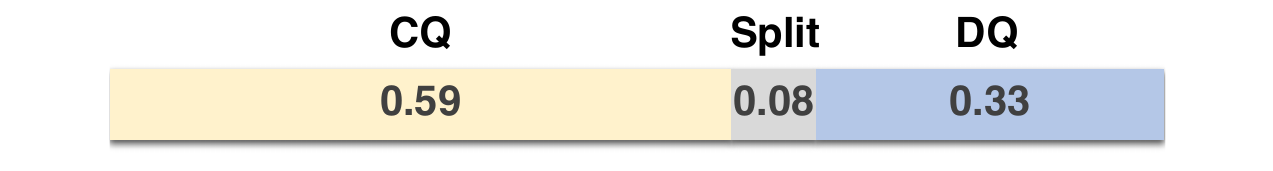}}\caption{
% \textcolor{red}{
Percentage of questions where the majority of people preferred ``CQ'', ``DQ'' and ``Split'', respectively. ``Split'' denotes that there was no majority response.
% }
}
\label{fig:preference}
\vspace*{-0.3cm}
\end{figure}
 
% \subsection{Preference test against DQ}
\paragraph{Setup} 
% We formulate preference tests to examine the user preferences between CQ and DQ.
% In the test, w
We randomly sampled 100 AQs from the development set. Then, for each AQ, we asked three annotators to show their preferences for ``CQ'', ``DQ'', or ``Equal'', along with the rationale.
That is, given an AQ, we ask people to compare ``being presented with answers to all possible interpretations (DQs) of the AQ" vs ``first answering a CQ and then being presented with an answer fitting their intention."
We then report the majority preference for each of the questions.
% We then report the majority preference for each question to prevent an annotator from having too much influence on the outcome.
% \textcolor{red}{The majority vote among the three individuals was determined and reported as the final preference.}
Please refer to Appendix~\ref{app:human_evaluation} for more details.

\paragraph{Results and Analysis}
% \textcolor{red}{
Figure~\ref{fig:preference} demonstrates that answering AQs using CQ is preferred over DQ. 
The prominent reasons stated by annotators for favoring ``CQ'' are its ease of use, conciseness, interactivity, and ability to provide clear guidance.
Conversely, annotators who preferred  ``DQ'' mentioned its advantage as being more specific and clearer in addressing the given question.
% These dominant reasons underscore the contrasting attributes of ``CQ'' and ``DQ'' that influenced the preferences of the annotators.

% Additionally, we specifically analyze instances where unanimity is observed among the three annotators.
Note, CQ was unanimously preferred 23 times, and DQ, 5 times.
Also, unanimity in favor of CQ was observed across AQs regardless of the number of interpretations---or options---whereas unanimity in favor of DQ only occurred for AQs with up to three interpretations. 
In other words, CQ can be preferred regardless of the number of interpretations, while DQ is not preferred when many interpretations are possible. 
This is intuitive given that more interpretations result in more text for people to process for the DQ-based response to AQ.
% These robust findings strongly support our initial motivation, emphasizing that CQ is better suited for real-world scenarios, particularly when multiple interpretations are present.
% }
% The results presented in Table~\ref{tab:ablation} demonstrate that the quality of generated CQs varies significantly when used to disambiguate AQs for a neural reader model in two realistic scenarios, namely ``No-answers" and ``Predicted answers".
% \section{RQ2: Do Plausible Answers to AQ Help in the Pipeline of Tasks?}
% \section{\textcolor{red}{Experiments}}
\section{Experiment 2: Handling AQ with CQ}
\label{sec:experiments}

% \textcolor{red}{Our proposed approach to \textit{first ask a CQ and present a single answer well-aligned with the user's intention} and the existing approach to \textit{presenting answers to all DQs}~\cite{min-etal-2020-ambigqa, stelmakh2022asqa} both seek to solve the issue of AQs in ODQA. However, since the problems are formulated differently, the performances of the models are not directly comparable. 
% Thus, rather than making comparisons to models for answering DQs, we focus on designing the state-of-the-art model for answering AQs using CQs.}

% \textcolor{red}{
% Our main focus is on designing a most effective model for answering AQs using CQs.
% To achieve this, we set up the experiments to compare two variations of utilizing the QA model: (1) initially running the QA model on the AQ and incorporating the predicted answers as input (the \textit{Predicted Answers for AQ} case) and (2) not running the QA model on AQ (the \textit{No Answers for AQ} case).
% \footnote{these variations were established based on the observation that an existing approach generated DQs using predicted answers. We aim to investigate whether this approach of predicting plausible answers before generating CQs could also be effective in the CQs framework}
% Given that running a QA model is an expensive step in ODQA, such experiments will provide insight into how to put together the ODQA pipeline with CQs.
% }

% \textcolor{red}{
Given that the CQs are preferred over DQs, 
% we now seek to achieve the best performance. In particular, we 
we now
study the question: \textit{Do predicted answers for AQ help improved the end-to-end performance of the CQ-based approach?} Since predicted answers for AQ have been shown to be helpful for previous DQ-based approaches~~\cite{min-etal-2020-ambigqa, gao2021answering}, we want to verify if they are also helpful for the CQ-based approach.\footnote{Note, while both CQ-based and DQ-based approaches seek to solve the issue of AQs in ODQA, the performances of the models are not directly comparable. This is because the problems are formulated differently---the former generates a CQ and an answer aligned with the user's intention, whereas the latter generates DQs and answers for them.}
For this, we experiment with two settings: 
\begin{enumerate}
\setlength\itemsep{0em}
    \vspace{-2mm}
    \item \textit{Predicted Answers for AQ:} running a QA model on the AQ and incorporating the predicted answers as input to the subsequent tasks 
    % \vspace{-2mm}
    \item \textit{No Answers for AQ:} not predicting answers to the AQ, and thus not using them in the subsequent tasks
\end{enumerate}
% (1) running a QA model on the AQ and incorporating the predicted answers as input to the subsequent tasks (the \textit{Predicted Answers for AQ} case) and (2) not predicting answers to the AQ, and thus not using them in the subsequent tasks (the \textit{No Answers for AQ} case).
% } 

In the remainder of the section, we present the experimental setup and results for each task. Please refer to Appendix~\ref{app:train_detail} for more details.

\subsection{Task 1: Ambiguity Detection}
\paragraph{Setup}
Since our dataset consists only of AQs, i.e., questions with multiple interpretations and answers, we combine it with unambiguous questions, i.e., questions with a single interpretation and answer, from \textsc{AmbigNQ} for this task. 

For \textit{No Answers for AQ} case, we use the \textsc{BERT$_\text{base}$} model \cite{Devlin2019BERTPO} with a simple fully connected layer on top for the binary classification task.
The model is trained on the combined dataset for 96 epochs.
The model also takes in a prompt of the form ``\textit{question [SEP] relevant\_passages}'' as input and outputs ``Ambiguous'' or ``Unambiguous''.

For \textit{Predicted Answers for AQ} case, we use BART-based model called \textsc{SpanSeqGen}, 
% \textcolor{red}{
the best-performing model for predicting answers for AQ by~\citet{min-etal-2020-ambigqa}, and finetuned it
% } 
on the \textsc{AmbigNQ} dataset.
This model takes in a prompt of the form ``\textit{question [SEP] relevant\_passages}'' as input and predicts all plausible answers.
We classify a question as ``Ambiguous" if the model outputs more than one plausible answer and ``Unambiguous,'' otherwise.

\begin{table}[]
\centering
\small
%\resizebox{\columnwidth}{!}{%
\begin{tabular}{lcccc}
\hline
%Classification by     & Acc.  & Pre. & Rec.  & F1   \\ \hline
Input in addition to AQ     & Acc.  & Pre. & Rec.  & F1   \\ \hline
No Answers for AQ   & \textbf{63.9} & \textbf{61.9} & \textbf{60.7} & \textbf{61.3} \\
Predicted Answers for AQ & 56.5 & 59.7 & 24.1 & 34.3 \\ \hline
\end{tabular}%
%}
\caption{
Evaluation results for the Ambiguity Detection task. 
The \textit{No Answers} case uses \textsc{BERT-base-cased} to determine whether a given question is ambiguous or not. 
% This case does not predict any answers but only progresses binary classification. 
The \textit{Predicted Answers} case makes use of answers predicted by \textsc{SpanSeqGen} and classifies the question as unambiguous only if exactly one answer is predicted.
}
\label{tab:AD_eval}
\vspace*{-0.3cm}
\end{table}

% \begin{table}[]
% \centering
% \resizebox{\columnwidth}{!}{%
% \begin{tabular}{c|cccc}
% \hline
% Model        & Acc  & Prec        & Rec         & F1          \\ \hline
% Span-Seq-Gen & 56.5 & 59.7 & 24.1 & 34.3  \\
% BERT         & 63.9 & 61.9 & 60.7 & 61.3 \\ \hline
% \end{tabular}%
% }
% \caption{
% Evaluation results for the Ambiguity Detection task. 
% }
% \label{tab:AD_eval}
% \end{table}
\begin{table*}[]
\renewcommand{\arraystretch}{1.0}
\centering
\small
%\resizebox{\textwidth}{!}{%
\begin{tabular}{lcccccccc}
\hline
\multirow{3}{*}{Input in addition to AQ and RPs} & \multicolumn{2}{c}{CQ} & \multicolumn{2}{c}{Category} & \multicolumn{4}{c}{Options} \\ \cmidrule(lr){2-3} \cmidrule(lr){4-5} \cmidrule(lr){6-9}
                    & BLEU-4 & \textsc{Bertscore} & EM   & BLEU-1 & Pre. & Rec. & F1   & Avg. \# \\ \hline
No Answers for AQ       & \textbf{7.9} &\textbf{88.9}   & 20.2 & \textbf{47.3}   & \textbf{37.4} & 18.2   & 24.5 & 2.0    \\
Predicted Answers for AQ  & \textbf{7.9} & \textbf{88.9}   & \textbf{22.8} & 44.0   & 36.9 & \textbf{19.0}   & \textbf{25.1} & 2.0    \\
\hline
Ground Truth Answers for AQ    & 15.4 &89.6   & 25.2 & 46.9   & 34.3 & 34.4   & 34.3 & 3.7    \\ \hline
\end{tabular}%
%}
\caption{
    Evaluation results for the Clarification Question (CQ) Generation task, where generated CQs are compared against the reference CQs. Each CQ was generated from an ambiguous question (AQ), relevant passages (RPs), and either \textit{No Answers}, \textit{Predicted Answers}, or \textit{Ground Truth Answers} for the AQ. The \textit{ground truth answers} case represents an ideal scenario in which the QA system perfectly identifies all possible answers for the AQ.
    }
    \label{tab:CQG_eval}
    % \vspace*{-0.1cm}
\end{table*}

\begin{table*}[]
\renewcommand{\arraystretch}{1.0}
\centering
\small
%\resizebox{\textwidth}{!}{%
\begin{tabular}{lcccccccc}
\hline
\multirow{3}{*}{CQ used to clarify the AQ}  &
  \multicolumn{4}{c}{NQ-pretrained BART} &
  \multicolumn{4}{c}{CQ-finetuned BART} \\ \cmidrule(lr){2-5} \cmidrule(lr){6-9}
 &
  \multicolumn{1}{c}{Pre.} &
  \multicolumn{1}{c}{Rec.} &
  \multicolumn{1}{c}{F1} &
  \multicolumn{1}{c}{\# Ans.} &
  Pre. &
  \multicolumn{1}{c}{Rec.} &
  \multicolumn{1}{c}{F1} &
  \multicolumn{1}{c}{\# Ans.} \\ \hline
% AQ revised by &
%   \multicolumn{1}{l}{} &
%   \multicolumn{1}{l}{} &
%   \multicolumn{1}{l}{} &
%   \multicolumn{1}{l}{} &
%   \multicolumn{1}{l}{} &
%   \multicolumn{1}{l}{} &
%   \multicolumn{1}{l}{} &
%   \multicolumn{1}{l}{} \\
\multicolumn{1}{l}{CQ generated with No Answers for AQ} &
  \multicolumn{1}{l}{47.9} &
  \multicolumn{1}{l}{25.2} &
  \multicolumn{1}{l}{33.0} &
  1.5 &
  54.4 &
  31.1 &
  39.6 &
  1.6 \\
\multicolumn{1}{l}{CQ generated with Predicted Answers for AQ} &
  \textbf{49.6} &
  \textbf{26.2} &
  \textbf{34.3} &
  1.5 &
  \textbf{55.4} &
  \textbf{32.0} &
  \textbf{40.5} &
  1.6 \\
  \hline
\multicolumn{1}{l}{CQ generated with Ground Truth Answers for AQ} &
  39.7 &
  37.5 &
  38.6 &
  2.0 &
  47.5 &
  49.5 &
  48.5 &
  2.5 \\

\multicolumn{1}{l}{Ground Truth CQ} &
  47.5 &
  39.8 &
  43.3 &
  2.0 &
  58.0 &
  53.8 &
  55.8 &
  2.5 \\
% Reference CQ &
%   47.5 &
%   39.8 &
%   43.3 &
%   3.3 &
%   58.0 &
%   53.8 &
%   55.8 &
%   3.3 \\ 
  \hline
\end{tabular}%
%}
\caption{
Evaluation results for the Clarification-based QA task. Answers found by a QA model for the AQs clarified with CQs are compared against the ground truth answers for the AQs.
% The Ground Truth Answers case is an ideal scenario in which the QA system perfectly identifies all possible answers for the AQ, based on which a CQ used to clarify the AQ is generated.
Three variations of model-generated CQs, derived from the CQ Generation task, are used to clarify the AQs.
The \textit{Ground Truth CQ} case is an ideal scenario in which Ground Truth CQs are used to clarify the AQs.
The \# Ans. is the average number of unique answers predicted for each AQ. 
%The \textit{Reference CQ} is another upper-bound case using the LLM-generated and human-edited CQ.
%To compute the scores for a given AQ, we compared the set of all possible answers (ground truth answers) to the set of answers found for the AQ clarified with the given CQ. (One answer was found assuming one of the options were chosen by the user. This process was repeated for all options to construct the set of answers.) Sets of answers were compared, because options in generated CQs often do not align well with those in the respective reference CQs. And even when the two sets of options do not align well, each set can still correctly represent different interpretations of the given AQ, and thus cannot be considered incorrect. (See Section~\ref{sec:eval_metrics} for more details.)
%The method of revising AQ with CQ is shown in Table~\ref{tab:cq_examples}. 
}
\label{tab:CQA_eval}
\vspace*{-0.3cm}
\end{table*}

\paragraph{Results and Analysis}
Table~\ref{tab:AD_eval} summarizes the result of ambiguity detection of two models \textsc{BERT$_\text{base}$} (\textit{No Answers}) and \textsc{SpanSeqGen} (\textit{Predicted Answers}). 
\textsc{SpanSeqGen} exhibits a similar precision as \textsc{BERT$_\text{base}$} (59.7 vs 61.9) but a significantly lower recall (24.1 vs 60.7).
This is because most questions are classified as ``Unambiguous.'' since the average number of answers generated by \textsc{SpanSeqGen} is 1.24. Consequently, this results in a much higher precision when compared to the recall for the same case.
This result indicates that classifying AQs by predicting all plausible answers is a challenging task for the Seq2Seq model.
% These results underscore the challenges associated with classifying AQ by predicting all plausible answers.
% \todo{Too short.. perhaps discuss a specific example that was correctly classified in the No Answers case, but not in the Predicted Answers case?}

\subsection{Task 2: Clarification Question Generation}
\label{sec:cqg_exp}
\paragraph{Setup}
For this task, we only use ground truth AQs to isolate the task from ambiguity detection. Please refer to Section~\ref{sec:end2end} for experiments in which errors do propagate from one task to the next.
% In other words, we used  with the train/validation/test split as described in Table~\ref{tab:data_statistics}.

% \textcolor{red}{
For \textit{No Answers for AQ}
% }
, we first train a BART$_\text{large}$
% \footnote{this choice was motivated by the fact that \textsc{SpanSeqGen} and the DQ generation model in previous research~\cite{min-etal-2020-ambigqa} are based on BART} 
model for 18 epochs, that takes ``\textit{AQ [SEP] relevant\_passages}'' as input and generates CQ as output. 
During inference, this model was used with a prompt of the form ``\textit{AQ [SEP] relevant\_passages}''.

% We first fine-tuned BART$_\text{large}$\footnote{\url{https://huggingface.co/facebook/bart-large}} for 18 epochs with the prompt ``\textit{AQ [SEP] relevant\_passages}" where.
% This model was used during inference with a prompt of the form ``\textit{AQ [SEP] relevant\_passages}'' for \textit{No Answers for AQ} case.

% To test whether predicting all plausible answers for AQ before generating a CQ enhances CQ generation, we also train a BART$_\text{large}$ model for 41 epochs, that takes ``\textit{AQ [SEP] possible\_answers\footnote{for training, the Ground Truth answers were used as possible\_ answers} [SEP] relevant\_passages}'' as input and generates CQ as output.
% During inference, we test two variations of adding \textit{possible\_answers} in the input: answers predicted by \textsc{SpanSeqGen} (\textit{Predicted Answers for AQ}) and the ground-truth answers (\textit{Ground Truth Answers for AQ}).

% \textcolor{red}{
For \textit{Predicted Answers for AQ}, we train another BART$_\text{large}$ model for 41 epochs, that takes ``\textit{AQ [SEP] possible\_answers
% \footnote{The ground truth answers were used for training.}
% \footnote{each possible\_answer are separated by [SEP]}
[SEP] relevant\_passages}'' as input and generates a CQ as output.
During inference, the model takes input with \textit{possible\_answers} as answers predicted by \textsc{SpanSeqGen}.
% }

% \textcolor{red}{
We also consider an additional setting, the \textit{Ground Truth Answers for AQ} case. This case is an ideal version of the \textit{Predicted Answers for AQ} case, where the ground truth answers are used as \textit{possible\_answers}. Hence, this case allows us to examine the effect of providing the correct set of plausible answers.

\paragraph{Results and Analysis}
The evaluation results of CQ generation with three variations of inputs are presented in Table~\ref{tab:CQG_eval}.
The results indicate that in the two realistic scenarios (\textit{No Answers for AQ}, \textit{Predicted Answers for AQ}), the quality of the generated CQs does not vary significantly in terms of the CQs themselves, the category, or the options.
This suggests that incorporating plausible answers as input to the CQ generation process does not improve the quality of the generated CQs in realistic scenarios.
However, when ground-truth answers are provided as input to the CQ generation process (\textit{Ground Truth Answers for AQ}), a significant improvement in the quality of the generated CQs is observed, but the quality is seemingly insufficient with a large room for improvement.
% This highlights the significant challenges inherent in the proposed task of CQ generation.

% To better understand the issue, we analyzed some predicted CQs that are evaluated as incorrect when compared against the reference CQs but are semantically correct. 
% As an example, for the question ``Who is the current chairman of African Union commission'' in Table~\ref{tab:changes_by_human}, the model generated ``Which chairman: 2017, or 2012?''. 
% The options ``\texttt{2017}'' and ``\texttt{2012}'' were classified as incorrect in CQ generation evaluation, but upon examination of the relevant passages, it was determined that the start term of the 4$^\text{th}$ and 3$^\text{rd}$ chairman of African Union are in fact 2017 and 2012, respectively. In other words, the two possible interpretations of the AQ could be expressed using the year as well as the term. However, it is difficult to align them given the differences in the surface representation, as well as the semantics.
% \textcolor{red}{
% To better understand the issue, we analyzed some predicted CQs that are evaluated as incorrect but semantically correct. 
In some cases, predicted CQs that are semantically correct were counted as incorrect.
For example, the model generated the CQ ``Which chairman: 2017 or 2012?'' for example 5 in Table~\ref{tab:changes_by_human}. 
Although deemed incorrect, a manual examination of relevant passages revealed the 4$^\text{th}$ and 3$^\text{rd}$ chairmen took office in 2017 and 2012, respectively. 
% }
This illustrates the challenge of directly comparing a predicted CQ and the respective reference CQ. Thus, the absolute score in Table~\ref{tab:CQG_eval} may not be as meaningful as the relative scores. Also, evaluating CQs in a downstream task may be necessary to better assess the qualities of the CQs, which we do in the clarification-based QA task.
% Thus, in addition to these intrinsic evaluation metrics, we also evaluate the quality of CQs through an extrinsic metric, which measures the effectiveness of the generated CQs in downstream tasks.

\subsection{Task 3: Clarification-based QA}
\label{sec:cbqa_exp}
\paragraph{Setup}
% \textcolor{red}{
We use NQ-pretrained BART$_\text{large}$ for the reader model which was trained on Natural Questions (NQ) dataset \cite{kwiatkowski-etal-2019-natural}.
% }
The model takes in an \textit{AQ clarified by CQ}---which is the concatenation of AQ, category, and option---and reranked relevant passages as input and predicts an answer for \textit{AQ clarified by CQ}. 
(See Section~\ref{sec:cbqa} for more details on \textit{AQ clarified by CQ})

In addition to the NQ-pretrained model, we also finetuned the NQ-pretrained reader model (CQ-finetuned BART) on our proposed dataset for 8 epochs.
During finetuning, the model also takes in an \textit{AQ clarified by CQ} as input.
The target label is the corresponding answer for each option.

During the inference, we employed three variations of model-generated CQs, derived from Section~\ref{sec:cqg_exp} CQ Generation task. Moreover, we consider an ideal scenario wherein the Ground Truth CQ is available and used to clarify the AQ.

\paragraph{Results and Analysis}
% Table~\ref{tab:res_extrinsic} shows that DQs revised by predicted CQs make the BART\footnote{BART$_\text{large}$ (BART) fine-tuned on \textsc{NQ} and its further fine-tuned one (BART-ft) on DQs and their answers in \textsc{AmbigNQ}.} recall more answers comparing to AQs ($20.9{>}17.5$).
The evaluation results of clarification-based QA using four variations of input and different reader models are presented in Table~\ref{tab:CQA_eval}.
% \textcolor{red}{
Two ideal settings (\textit{CQ generated with Ground Truth Answers for AQ} and \textit{Ground Truth CQ}) exhibit lower precision scores.
% the two realistic settings (\textit{No Answers for AQ}, \textit{Predicted Answers for AQ}) exhibit higher precision scores.
% }
% three variations of input tend to display similar precision tendencies in terms of orders.
On the other hand, they outperform the other two variations (\textit{CQ generated with No Answers for AQ} and \textit{CQ generated with Predicted Answers for AQ}) in terms of recall, resulting in higher F1 scores, as well.
One reason for this is that the \textit{CQs generated by  Ground Truth Answers for AQ} and \textit{Ground Truth CQs} contain more options (1.5 more on average) which leads to predicting more answers than the other two variations, resulting in higher recall and lower precision scores.

The average numbers of options in Table~\ref{tab:CQG_eval} and those of unique answers in Table~\ref{tab:CQA_eval} indicate that both NQ-pretrained BART and CQ-finetuned BART struggle to generate distinct answers for distinct options.
For instance, in the \textit{CQ generated with Ground Truth Answers for AQ} case, where the average number of options for CQs is 3.7, 
% we would expect the reader model to generate an average of 3.7 distinct answers for each AQ. 
% This expectation arises from the fact that  each option should result in a distinct answer. 
% However, 
only 2.5 distinct answers were generated for each AQ.
In other words, both models tend to produce the same answer for the given AQ even if the specified options are different.
% }
% Specifically, in the upper-bound \textit{Ground Truth Answers} case, we find that the average number of answers generated by the reader model, 2.5, is smaller than the average number of options, 3.7, as shown in Table~\ref{tab:CQG_eval}.
% \textcolor{red}{
This phenomenon, referred to as the ``collapse" of the models has also been reported in previous studies such as \cite{zhang-choi-2021-situatedqa}. It suggests that deep learning models can be insensitive to subtle differences in the input---when different options are chosen for the same AQ, the input would be identical except for the option.
% }

\subsection{End-to-End}
\label{sec:end2end}
\paragraph{Setup}
% \textcolor{red}{
% Based on the results from each task, we design 
We now conduct experiments to 
check RQ2---whether predicted answers for AQ help improve the CQ-based approach to handle AQ end-to-end.
% identify the most effective setting for answering AQs using CQs.
We consider four combinations of setting for ambiguity detection and CQ generation: 
\begin{enumerate}
    \vspace{-2mm}
    \setlength\itemsep{0em}
    \item \textit{Pred Answers--Pred Answers}: running a QA model on the AQ and incorporating the predicted answers in both tasks 
    \item \textit{No answers--No Answers:} not running the QA model on AQ 
    \item \textit{Pred Answers--No Answers:} running the QA model on the AQ but using the predicted answers as input for ambiguity detection only
    \item \textit{No Answers--Pred Answers:} running the QA model on the AQ but using the predicted answers as input for CQ generation only
\end{enumerate}

% (1) running a QA model on the AQ and incorporating the predicted answers for both tasks (\textit{Pred Answers--Pred Answers}), 
% (2) not running the QA model on AQ (\textit{No answers--No Answers}), 
% (3) running the QA model on the AQ but only utilizing the predicted answers for ambiguity detection (\textit{Pred Answers--No Answers}), and (4) 
% running the QA model on the AQ but only utilizing the predicted answers for CQ generation (\textit{No Answers--Pred Answers}).
% }
% We design an end-to-end experiment to identify the best time to run the QA model: before ambiguity detection (\textit{Pred Answers--Pred Answers}), between ambiguity detection and CQ generation (\textit{No answers--Pred Answers}), or after CQ generation (\textit{No answers--No answers})? We also test the possibility of running the QA model before ambiguity detection, but not using the predicted answers during CQ generation (\textit{Pred Answers--No Answers}) for the sake of completeness.
% We report the end-to-end experiment to see the combined results for our entire task pipeline.
% In our experimental setup for the end-to-end process, we evaluate four different scenarios, which are a combination of two realistic settings (Classification by \textit{No Answers}, \textit{Predicted Answers}) in AD and two realistic settings (\textit{No Answers for AQ}, \textit{Predicted Answers for AQ}) in CQ generation. 
The end-to-end performances are measured at the end of the pipeline, i.e., clarification-based QA.

\begin{table}[]
\centering
\small
\begin{tabular}{cccccc}
\hline
Ambig. Detect.                        & CQ Gen.  & Pre. & Rec. & F1    \\ \hline
\multirow{2}{*}{No Answers}   & No Answers   & \textbf{43.2} & \textbf{19.9} & \textbf{27.3}   \\
                          & Pred Answers & 42.8 & 19.6 & 26.9   \\ \hline
\multirow{2}{*}{Pred Answers} & No Answers   & 22.5 & 8.3  & 12.1   \\
                          & Pred Answers & 24.7 & 9.0  & 13.1 \\ \hline
\end{tabular}
\caption{
End-to-end Evaluation Results: The performances are measured at the end of the pipeline, i.e., clarification-based QA. 
% Ambiguity Detection (AD) and Clarification Question (CQ) Generation. 
% In AD, BERT and \textsc{SpanSeqGen} were used for \textit{No Answers} and \textit{Pred Answers} cases, respectively. The reader model is finetuned on CQ.
}
\label{tab:ablation}
\vspace*{-0.3cm}
\end{table}

% \begin{table}[]
% \centering
% \small
% %\resizebox{\columnwidth}{!}{%
% \begin{tabular}{ccccc}
% \hline
% CQ Gen.  & Pre. & Rec. & F1   & #Ans \\ \hline
% No-Ans   & 43.2 & 19.9 & 27.3 & 2.0  \\ \hline
% Pred-Ans & 24.1 & 8.9  & 13.0 & 2.0  \\ \hline
% \end{tabular}%
% %}
% \caption{
% Ablation on the end-to-end process: Ambiguity Detection (AD) and Clarification Question (CQ) Generation. For AD, BERT and \textsc{SpanSeqGen} were used for \textit{No-Ans} and \textit{Pred-Ans} cases, respectively. The reader model is finetuned on CQ.
% }
% \label{tab:ablation}
% \end{table}
% \begin{table}[]
% \centering
% \small
% %\resizebox{\columnwidth}{!}{%
% \begin{tabular}{llcccc}
% \hline
% CQ Gen. & AQ & Pre.  & Rec.  & F1   & \# Ans. \\ \hline
% \multirow{2}{*}{No-Ans}   & True AQ & 54.4 & 31.1 & 39.6 & 2.0    \\
%                               & Pred AQ    & 43.2 & 19.9 & 27.3 & 2.0    \\ \hline
% \multirow{2}{*}{Pred-Ans} & True AQ & 55.8 & 32.8 & 41.3 & 2.0    \\
%                               & Pred AQ   & 24.1 & 8.9  & 13.0 & 2.0    \\ \hline
% \end{tabular}%
% %}
% \caption{
% Ablation on the end-to-end process: Ambiguity Detection (AD) and Clarification Question (CQ) Generation. For AD, BERT and \textsc{SpanSeqGen} were used for \textit{No-Ans} and \textit{Pred-Ans} cases, respectively. The reader model is finetuned on CQ.\todo{
% Shouldn't the rows be:
% No-Ans / No-Ans;
% No-Ans / Pred-Ans;
% Pred-Ans / No-Ans;
% Pred-Ans / Pred-Ans;
% for columns AD and CQ Gen?
% }
% }
% \label{tab:ablation}
% \end{table}

\paragraph{Results and Analysis}
% \todo{Provide a description of what this experiment is about. Also, it may make sense to restructure Section 5 by task: 5.1 Ambiguity detection (experimental setup & results), 5.2 CQ generation (experimental setup & results), 5.3 End-to-End (experimental setup & results). If you decide to restructure Section 5, you should do the same in Section 3: 3.1 Ambiguity detection (description & evaluation metric). 3.2 CQ generation (description & evaluation metric). If you do the restructuring, Subsection titles will be shorter, and it would be easier to readers to read. This is not recommended if experimental setup or task descriptions for CQ generation only makes sense right after reading them for ambiguity detection.}
As shown in Table~\ref{tab:ablation}, the use of the BERT model (\textit{No Answers}) for ambiguity detection and prompting without answers (\textit{No Answers}) in the input for CQ generation yields the highest F1 score of 27.3.
However, the combination of using the BERT model (\textit{No Answers}) for ambiguity detection and utilizing predicted answers by the \textsc{SpanSeqGen} model (\textit{Pred Answers}) in the input for CQ generation resulted in an F1 score 0.4 lower than the best combination.
% \textcolor{red}{
Note, the \textit{No Answers--Pred Answers} setting is not only (slightly) worse than the best approach, but is also inefficient as it requires running both BERT and \textsc{SpanSeqGen} models during inference.
% }

% \textcolor{red}{
\textit{No Answers--No Answers} and \textit{Pred Answers--Pred Answers} are the only settings in which only a single model is used for ambiguity detection and generating input for CQ generation. Among these, the quality of the generated CQs varies significantly.
% When considering only the efficient scenarios for inference, in which only a single model is used for ambiguity detection and generating input for CQ generation, namely the \textit{No Answers--No Answers} and \textit{Pred Answers--Pred Answers} scenarios, the quality of the generated CQ varies significantly.
% } 
More specifically, the results show that in the \textit{Pred Answers--Pred Answers} scenario, the poor performance of the ambiguity detection stage propagates to the remainder of the pipeline. 
This suggests that incorporating plausible answers as input to the CQ generation process prior to generating the CQs is not a desirable approach in the CQ framework.
Finally, the end-to-end performance of all four cases still has a large room for improvement, showing the challenging nature of CQ-based approach to handling AQs in ODQA, as well as the need for resources like \textsc{CAmbigNQ}.

\section{Conclusion}

We proposed a CQ-based approach to handle AQs in ODQA. 
% To this end, we presented \textsc{CAmbigNQ}, a dataset consisting of 5,653 ambiguous questions, each with relevant passages, possible answers, and a clarification question. The clarification questions were efficiently created by generating them using InstructGPT and manually revising them as necessary.
% Along with the dataset, we also defined a pipeline of tasks, designing appropriate evaluation metrics for it.
Along with presenting a new dataset, we defined a pipeline of tasks, designing appropriate evaluation metrics for it.
Experiments show the promising, yet challenging nature of the tasks.
% We presented the first study of generating clarification questions for open-domain question answering.
% We introduced \textsc{CAmbigNQ}, a dataset consisting of 5,653 ambiguous questions and clarification questions, constructed through generation by InstructGPT and revision by human editors.
% We defined a pipeline of tasks for handling ambiguity in ODQA, designing appropriate evaluation metrics for it.
% We achieved 40.5 F1 on clarification-based QA and 27.3 F1 on end-to-end experiments, forming strong baselines for future work.
We hope that our dataset will serve as a valuable resource for research in this area and support the development of more user-friendly QA systems.
% \textcolor{red}{
% We leave for future work \textit{Predicted Answers for AQ} experiments with other models designed for the DQ-based approach~\cite{gao2021answering}.
% Finally, it is worth noting that there is potential for improvement in the \textit{Predicted Answers for AQ} cases for each subsequent task. This potential improvement can be realized by utilizing advanced models for plausible answer prediction, such as Refuel~\cite{gao2020answering}. 
% However, this specific comparison is deferred for future work.
% }
\section*{Limitations}
As shown in our results, both clarification question generation and clarification-based question answering evaluations can still underestimate the performance of the generated clarification questions due to various factors.
One reason is that the reference clarification questions are one of many possible answers, not the only correct answer.
Another reason is that the intrinsic evaluation, which depends on the overlap between the texts, may not properly handle semantically correct predictions.
Additionally, the extrinsic QA model for clarification-based question answering may fail to perform reasoning.
These limitations highlight the need for further research in the field to improve evaluation methods for clarification question generation tasks.

% As indicated in our results, both intrinsic and extrinsic evaluation can underestimate the generated clarification questions due to various factors.
% First, reference clarification questions are just one of many possible answers, not the only correct answer.
% Second, the intrinsic evaluation which depends on the overlap between texts mishandles semantically correct predictions.
% Third, the extrinsic QA model fails to do reasoning.
% One may concern that our auto-converted clarification questions are not human-written questions. 
% In fact, we tried to verify the quality of the dataset by using HIT. 
% However, most of the annotators marked that our auto-converted CQs already satisfy the requirements. 
% This indicates that our CQs are already similar to human-written CQs. 
% auto-conversion reliability. 
% 사실 HIT 를 해서 고치려고 했는데 사람들이 거의 고치지 않아서 human-annotated 를 얻는데 실패 했다.
% dataset 논문에 당연히 있어야 되는게 있어야됨. SituatedQA 보고 빠진게 머있는지 확인 NQ랑 

% extrinsic evaluation을 보장하지 못한다? 아직 CQ evaluation 관련해서 extrinsic eval 이 사람의 만족도와 correlation이 있다는 것을 알아내지 못했음

\section*{Ethics Statement}
Our proposed datasets will not pose any ethical problems as they have been constructed from the publicly available \textsc{AmbigNQ} \cite{min-etal-2020-ambigqa} dataset, which itself is derived from the Natural Questions dataset \cite{Kwiatkowski2019NaturalQA}.

Additionally, the use of the InstructGPT model for the generation of data was done by utilizing the official website of OpenAI\footnote{\url{https://openai.com/}}.
All models used in the experiments are from the publicly available website or Github.
While there is a possibility of bias or toxicity in the generated text, such issues are addressed through our human validation process.
Furthermore, the data annotators were fairly compensated for their work, and the details of payment can be found in Appendix~\ref{app:data_collection}.
\section*{Acknowledgements}
K. Jung is with ASRI, Seoul National University, Korea. This work has been financially supported by SNU-NAVER Hyperscale AI Center. This work was partly supported by Institute of Information \& communications Technology Planning \& Evaluation (IITP) grant funded by the Korea government(MSIT) [NO.2021-0-01343, Artificial Intelligence Graduate School Program (Seoul National University)]

\bibliography{ref}
\bibliographystyle{acl_natbib}
\appendix
\newpage
\section{Details of a partial match with the Hungarian algorithm}

% In this section, we adopt the following notations: the reference CQ contains the category $cat$ and the options $O = \{o_1, \dots, o_N\}$, and the predicted CQ $\hat{q_c}$ contains the category $\hat{cat}$ and the options $\hat{O} =\{\hat{o}_1, \dots, \hat{o}_M\}$, respectively, where $N$ and $M$ may differ.

\subsection{Alignment for Clarification Question Generation task}
\label{app:intrinsic}

% To find the optimal alignment between predicted options and reference options, we first calculate two overlap score matrices: one based on precision ($S_{N{\times}M}^{prec}$) and one based on recall ($S_{N{\times}M}^{rec}$).
% Each matrix contains the overlap score between a reference option $o_n$ and a predicted $\hat{o}_m$ in $S_{nm}$, based on either precision or recall.
% The elements of these matrices are defined as follows:
% \begin{equation}
% \begin{aligned}
% \textstyle{S_{nm}^{prec} = s^{prec}(o_n, \hat{o}_m)}
% \end{aligned}
% \end{equation}
% \begin{equation}
% \begin{aligned}
% \textstyle{S_{nm}^{rec} = s^{rec}(o_n, \hat{o}_m)}
% \end{aligned}
% \end{equation}
% % $S_{mn}^{prec}= max(s^{prec}(\hat{o}_m, o_1), ..., s^{prec}(\hat{o}_m, o_N))$ 
% % $S_{mn}^{rec   }= max(s^{rec   }(\hat{o}_1, o_n), ..., s^{rec   }(\hat{o}_M, o_1))$ 
% \paragraph{}
% \noindent where $s$ is a text similarity metric based on the text overlap defined as follows:
% \begin{equation}
% \begin{aligned}
% \resizebox{.85\hsize}{!}{$s^{prec}(o_n, \hat{o}_m)= len(LCS(o_n, \hat{o}_m)/len(\hat{o}_m)$}
% \end{aligned}
% \label{eq:sim_prec}
% \end{equation}
% \begin{equation}
% \begin{aligned}
% \resizebox{.85\hsize}{!}{$s^{rec   }(o_n, \hat{o}_m)= len(LCS(o_n, \hat{o}_m)/len({o_n})$}
% \end{aligned}
% \label{eq:sim_rec}
% \end{equation}

% \paragraph{}
The similarity function $sim(x, y)$ is defined as follows:
\begin{equation}
\begin{aligned}
\textstyle{sim(x, y) = len(LCS(x,y)) / len(x)}
\end{aligned}
\label{eq:sim_metric}
\end{equation}
In Equation~\ref{eq:prec} and ~\ref{eq:rec}, similarity scores between prediction and reference are calculated by dividing the length of the longest common subsequence by the length of the predicted option and reference option, respectively.

To match each predicted option to a corresponding reference option, the Hungarian algorithm~\cite{Kuhn1955TheHM}\footnote{The Hungarian algorithm, used for assigning tasks to workers in a one-to-one manner with the objective of minimizing the cost, is adapted in our study to maximize the cost by altering the setting.} is applied, and optimal correspondence mapping function $f_{i}$ for \textit{i}-th option is obtained.

\subsection{Alignment for Clarification-based Question Answering}
\label{app:extrinsic}
In the evaluation for CBQA, the reference answers are not a single string but rather a list of strings that may represent a single answer.
% For example, the reference answer set for the question ``Who is Michael Jordan?" may include the strings ``MJ", ``Michael Jefferey Jordan", and ``Jordan".
In this sense, the $\text{max}_{i}^{prec}$ and $\text{max}_{i}^{rec}$ for \textit{i}-th example is calculated differently as follows:

\begin{equation}
\begin{aligned}
% \resizebox{.85\hsize}{!}{\text{max}_{i}^{prec} = \sum_{p\in P_i} \max_j sim(p, f_{i}(p)_{j} )},
\textstyle{{\text{max}_{i}^{prec} = \sum_{p\in P_i} \max_j sim(p, f_{i}(p)_{j} )},}
\end{aligned}
\label{eq:max_prec}
\end{equation}

where, $f_{i}(p)_{j}$ is the single representation (e.g. ``MJ" for the example from Section~\ref{sec:cbqa}) of the reference answer set and prediction p is aligned to a single reference answer set consisting of total $J$ representations.

\begin{equation}
\begin{aligned}
% \resizebox{.85\hsize}{!}{\text{max}_{i}^{rec} = \sum_{r\in R_i} 
% \max_j sim(r_j, f_{i}^{-1}(r))},
\textstyle{{\text{max}_{i}^{rec} = \sum_{r\in R_i} 
\max_j sim(r_j, f_{i}^{-1}(r))},}
\end{aligned}
\label{max_rec}
\end{equation}

where, r is not the single string but a list of J strings that may represent a single answer. (e.g. [``MJ", ``Michael Jefferey Jordan'', ``Jordan''] for the example from Section~\ref{sec:cbqa}) and all representations in the list r are aligned to the same prediction by $f_{i}^{-1}(r)$.

% The overlap score matrices used in the evaluation for CBQA task $S_{N{\times}M}^{prec}$, $S_{N{\times}M}^{rec}$ are calculated differently from those used in the evaluation for CQ Generation task.
% In the evaluation for CBQA, the reference answers are not a single string but rather a list of strings that may represent a single answer.
% For example, the reference answer set for the question ``Who is Michael Jordan?" may include the strings ``MJ", ``Michael Jefferey Jordan", and ``Jordan".
% To calculate the overlap scores for the evaluation, the maximum overlap score between the predicted answer $\hat{a}_m$ for the AQ revised by CQ ($AQ, \hat{cat}: \hat{o}_m?$) and the reference answer set $A_n=\{a_{n1}, ... a_{nK}\}$ is taken as the element in the overlap score matrix $S_{nm}$.
% Each component of these matrices can be defined as follows:
% \begin{equation}
% \begin{aligned}
% \resizebox{.85\hsize}{!}{$S_{nm}^{prec}= max(s^{prec}(a_{n1}, \hat{a}_m), ..., s^{prec}(a_{nK}, \hat{a}_m))$}
% \end{aligned}
% \end{equation}
% \begin{equation}
% \begin{aligned}
% \resizebox{.85\hsize}{!}{$S_{nm}^{rec   }= max(s^{rec   }(a_{n1}, \hat{a}_m), ..., s^{rec   }(a_{nK}, \hat{a}_M))$}
% \end{aligned}
% \end{equation}
% %\paragraph{}
% \noindent where,
% similarity metrics $s^{prec}$ and $s^{rec}$ are the same as Equation~\ref{eq:sim_prec} and Equation~\ref{eq:sim_rec} respectively.

% The same procedure from the evaluation for CQ generation task is done on the overlap score matrices $S_{N{\times}M}^{prec}$, $S_{N{\times}M}^{rec}$.

\section{Details of data collection and dataset}
\label{app:data_collection}
\subsection{Details of data generation by LLM}
We use OpenAI (text-davinci-003) API model for the generation.
For the hyperparameters of InstructGPT, we limited the \textit{maximum length} of the output token to 100 and used 0.7 for \textit{temperature}.
We set \textit{top\_p} and \textit{n} to 1.
As mentioned in Section~\ref{sec:dataset}, the prompt is in the form of a concatenation of the six examples, instructions, target AQ, and target DQs.
Each example used for the prompt is described in Table~\ref{tab:fewshot}.

\subsection{Details of Manual Editing}
\begin{figure*}
        \centering
        \begin{subfigure}[b]{0.475\textwidth}
            \centering
            \includegraphics[width=\textwidth]{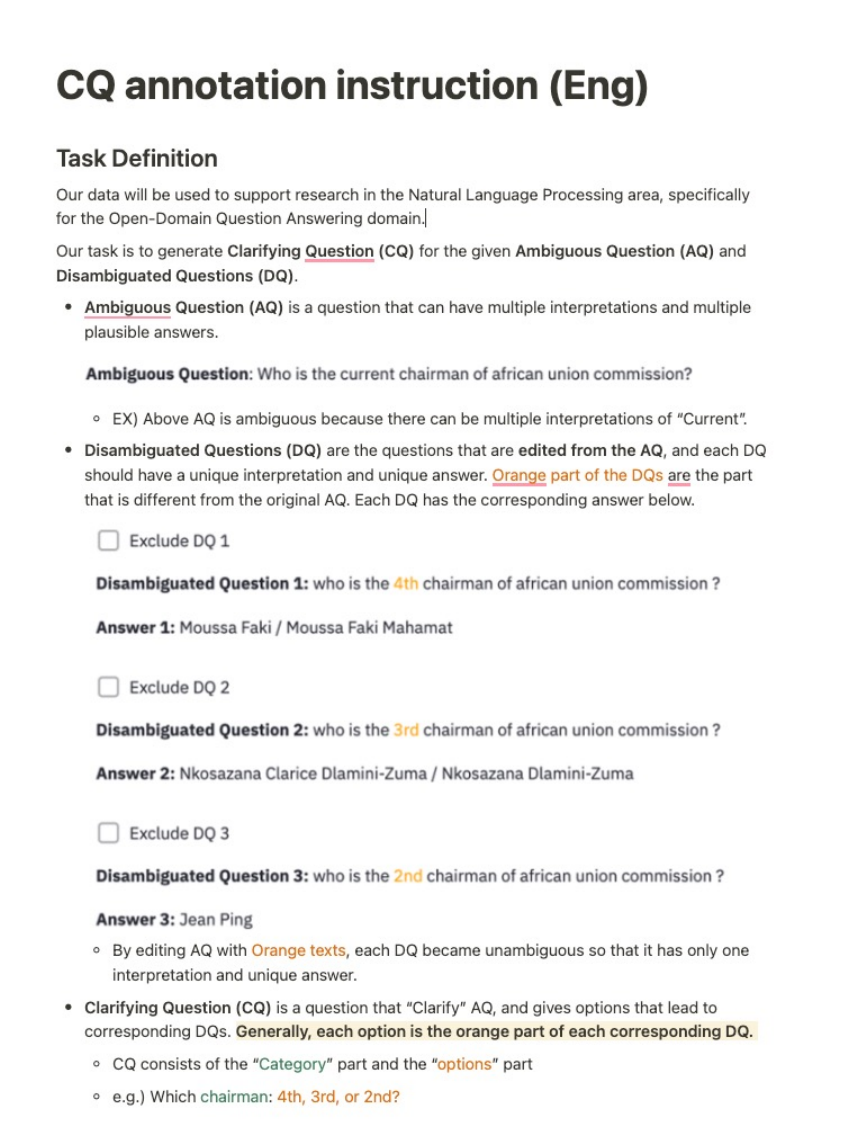}
            \caption[Network2]%
            {{\small CQ revision instruction page 1.}}    
            \label{fig:instruct1}
        \end{subfigure}
        \hfill
        \begin{subfigure}[b]{0.475\textwidth}  
            \centering 
            \includegraphics[width=\textwidth]{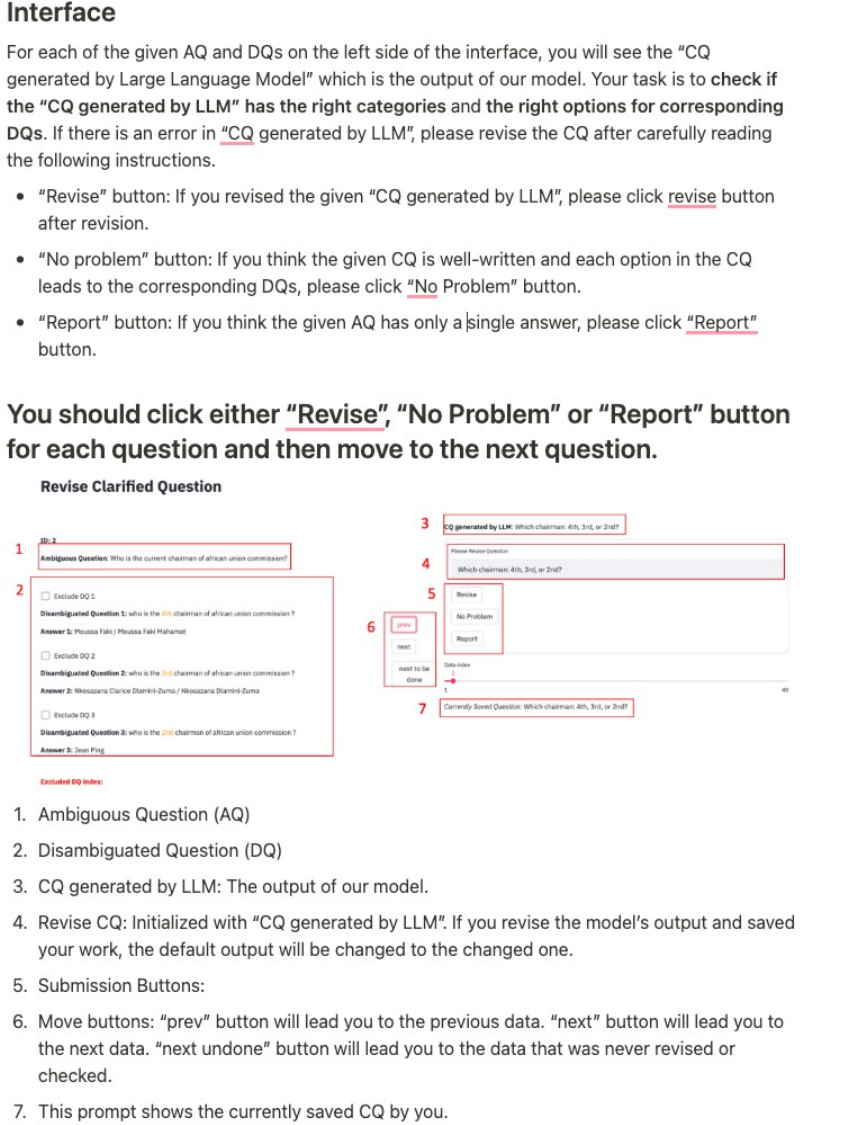}
            \caption[]%
            {{\small CQ revision instruction page 2.}}    
            \label{fig:instruct2}
        \end{subfigure}
        \vskip\baselineskip
        \begin{subfigure}[b]{0.475\textwidth}   
            \centering 
            \includegraphics[width=\textwidth]{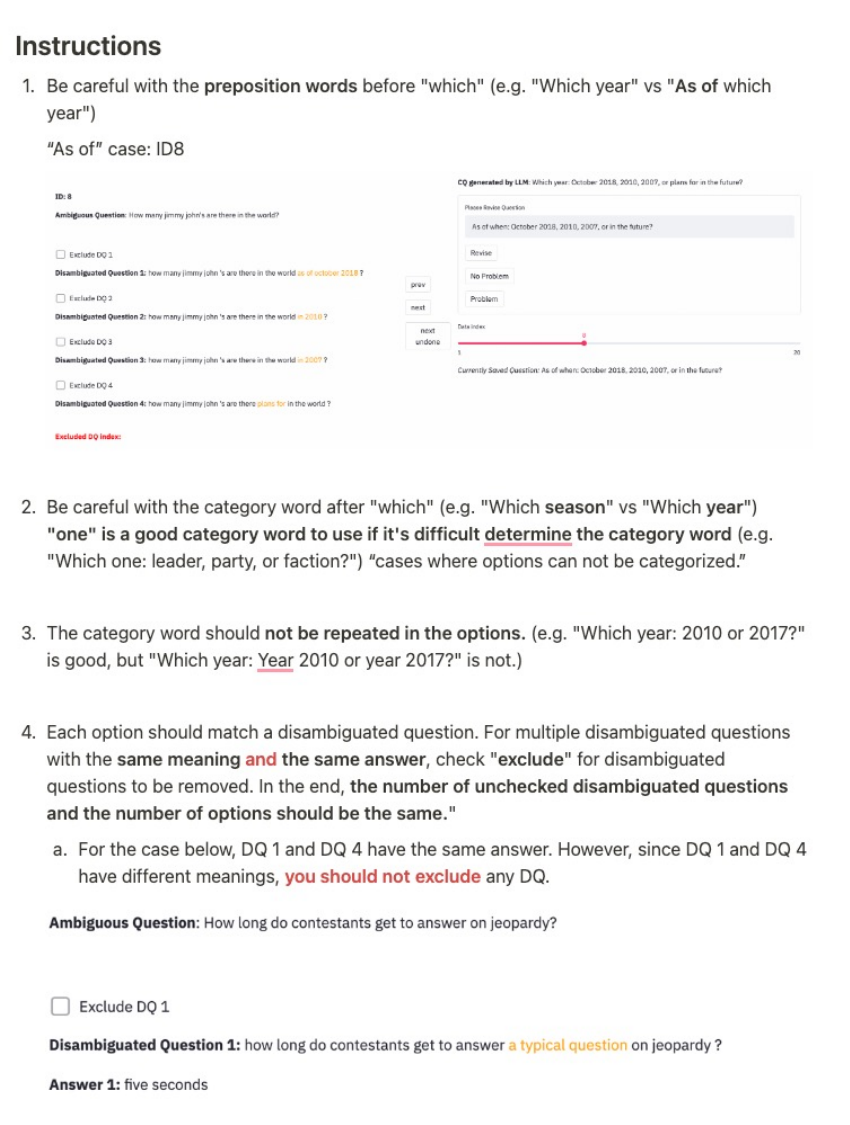}
            \caption[]%
            {{\small CQ revision instruction page 3.}}    
            \label{fig:instruct3}
        \end{subfigure}
        \hfill
        \begin{subfigure}[b]{0.475\textwidth}   
            \centering 
            \includegraphics[width=\textwidth]{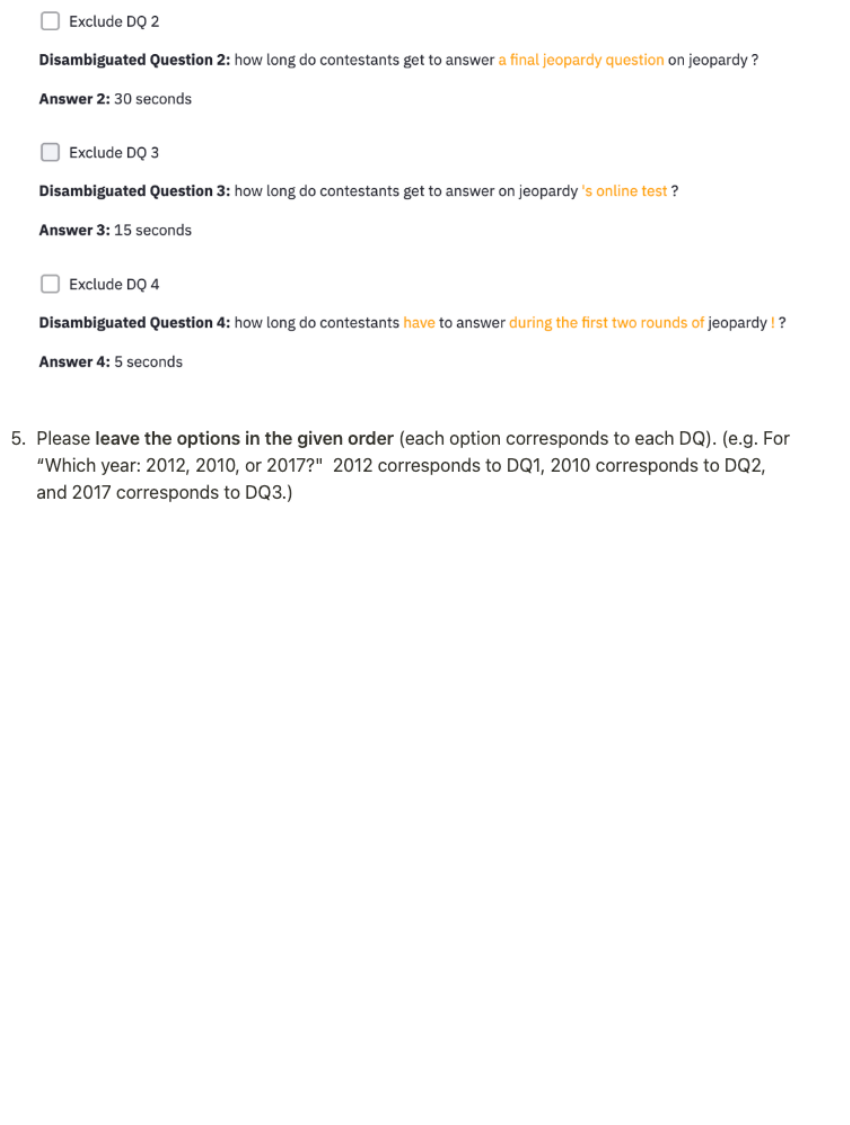}
            \caption[]%
            {{\small CQ revision instruction page 4.}}    
            \label{fig:instruct4}
        \end{subfigure}
        \caption[ The average and standard deviation of critical parameters ]
        {The instructions provided to the recruited annotators for CQ revision.} 
        \label{fig:instructions}
    \end{figure*}

\begin{figure*}
        \centering
        \begin{subfigure}[b]{0.9\textwidth}
            \centering
            \includegraphics[width=\textwidth]{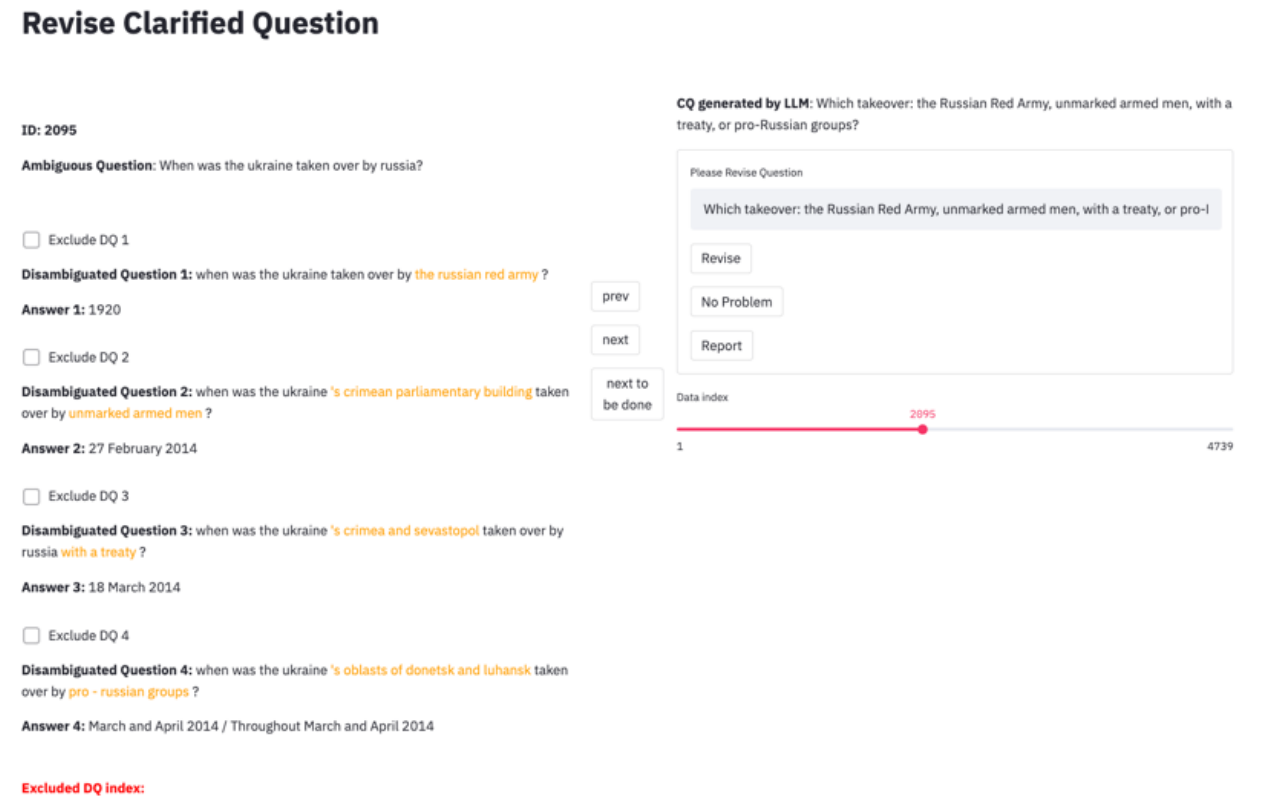}
            \caption[Network2]%
            {{\small Interface page for CQ revision example 1.}}    
            \label{fig:interface1}
        \end{subfigure}
        \hfill
        \begin{subfigure}[b]{0.9\textwidth}   
            \centering 
            \includegraphics[width=\textwidth]{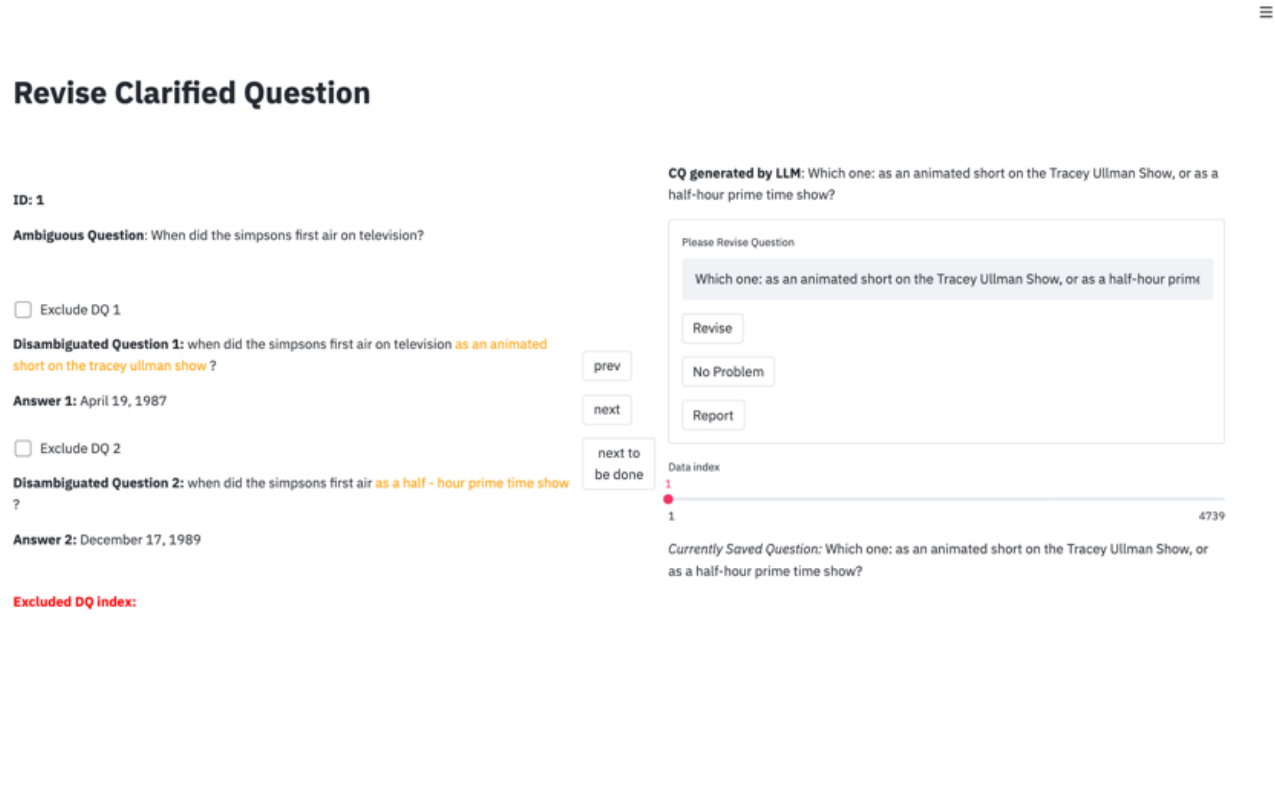}
            \caption[]%
            {{\small Interface page for CQ revision example 2.}}    
            \label{fig:interface2}
        \end{subfigure}
        \caption[ The average and standard deviation of critical parameters ]
        {Interface of qualification and manual editing stage for CQ revision.} 
        \label{fig:interfaces}
    \end{figure*}

Ten undergraduate students fluent in English were recruited through the university's online community, and seven of them successfully passed the qualification test.
The recruited annotators were provided with a detailed description of task definitions, instructions, and examples as shown in Figure~\ref{fig:instructions}, ~\ref{fig:interfaces}.
During the recruitment process, all applicants were informed that their annotations would be utilized for academic purposes and would be included in published papers.
This information was explicitly stated in the recruitment announcement and instructions to ensure transparency.
The annotators were then asked to review 25 examples that had been previously annotated by co-authors and revise 25 CQs generated by the InstrcutGPT model.

%Interface 
Seven annotators who passed the qualification stages were then selected to participate in the manual editing stages.
As shown in Figure~\ref{fig:instructions}, the annotators were provided with \textit{Ambiguous Question} and \textit{Disambiguated Questions} on the left side of the page.
To assist the annotation process, we used a process to identify the longest common subsequence between the AQ and DQs, a \textit{spaCy} constituency parser to identify different constituent parts of the DQs, and highlighted these parts.
On the right side of the page, the InstructGPT-generated CQ was provided, and the annotators were given the option to revise, pass (no revision), or report (single interpretation and answer) the given CQs.

% process
We used streamlit\footnote{\url{https://streamlit.io/}}, an open-source app framework for creating web apps for data science and machine learning, to construct the interface.
The InstructGPT-generated examples were divided into sets of 500 examples, and for quality control, we included 20 validation examples that were annotated by a co-author in each set.
The annotators were notified of the existence of the validation examples and asked to re-annotate the samples if the correct percentage of the correctly annotated validation examples did not meet a pre-determined threshold.

For the payment of the annotators, the co-authors first conducted annotations for an hour to estimate the average number of annotations that could be completed within an hour.
Based on this estimation, a rate of 0.15 dollars per example was established to ensure that the annotators would be paid at least 133\% of the minimum wage.

\subsection{Details of Inter-Annotator Agreement}
\label{app:IAA}
We conducted an evaluation of the output from two annotators and report the BLEU score and EM score. The BLEU score and EM score for the entire CQ are 65.8 and 39.3, respectively. When considering the category and options separately, the BLEU score for the category is 76.5 with an EM score of 56.8, while the BLEU score for options is 66.0 with an EM score of 63.9. All scores provided have been micro-averaged.

We conduct an analysis of instances in which there was disagreement between the two annotators. 
The primary cause of these disagreements can be attributed to variations in the specificity of categories, or the options provided. 
For example, in Figure~\ref{fig:amb_ex}, there were different opinions within the category of ``version'' and the first option ``young in series 2''.
The other annotator suggested alternatives such as ``version of Tom Riddle'' for the category and ``the young Tom Riddle in Harry Potter and the Chamber of Secrets'' for the first option. These alternatives are accurate in capturing the intended meaning, but they differ in their surface form.

\subsection{Details of human editing}
We provide a deeper analysis of human editing made on examples generated by InstructGPT, as shown in Table~\ref{tab:changes_by_human}. 
Specifically, for instances where human annotators made partial revisions, focusing on either the category or the options alone (referred to ``Category Only" and ``Options Only" in Table~\ref{tab:changes_by_human}), we compute the BLEU score.
Interestingly, both the ``Category Only" and ``Options Only" cases exhibit BLEU scores of 37.0 and 53.6, respectively.
% These scores suggest that the model's predictions align relatively close with the human annotators' perspectives, despite some disparities.
Additionally, in cases where the model generated invalid forms (referred to ``Whole Question'' in Table~\ref{tab:changes_by_human}), the BLEU scores between the model's predictions and the human revisions yield 36.7.
% This indicates that although the model struggles to capture the inherent format of a few-shot example, it still generates outputs that closely resemble natural language expressions.
It is important to note that while BLEU scores may not capture semantic similarity, they do provide valuable insights into the disparity between human-labeled data and model-generated data.

\subsection{Details of dataset}
\begin{figure*}[]
\centerline{\includegraphics[width=0.9\textwidth]{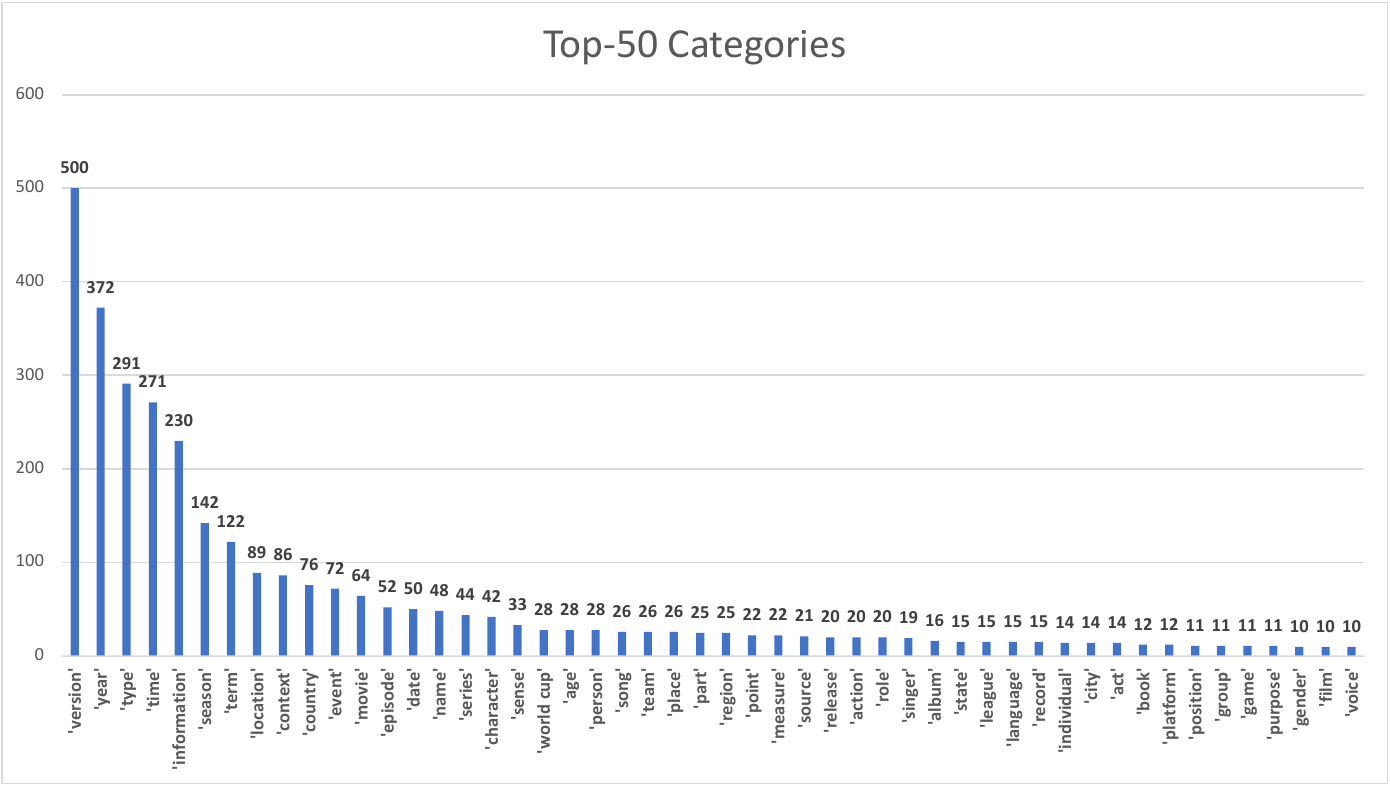}}
\caption{
Top-50 categories in \textsc{CAmbigNQ}.
}
\vspace{1cm}
\label{fig:category_distribution}
\end{figure*}

\begin{figure*}[t]
\centerline{\includegraphics[width=0.9\textwidth]{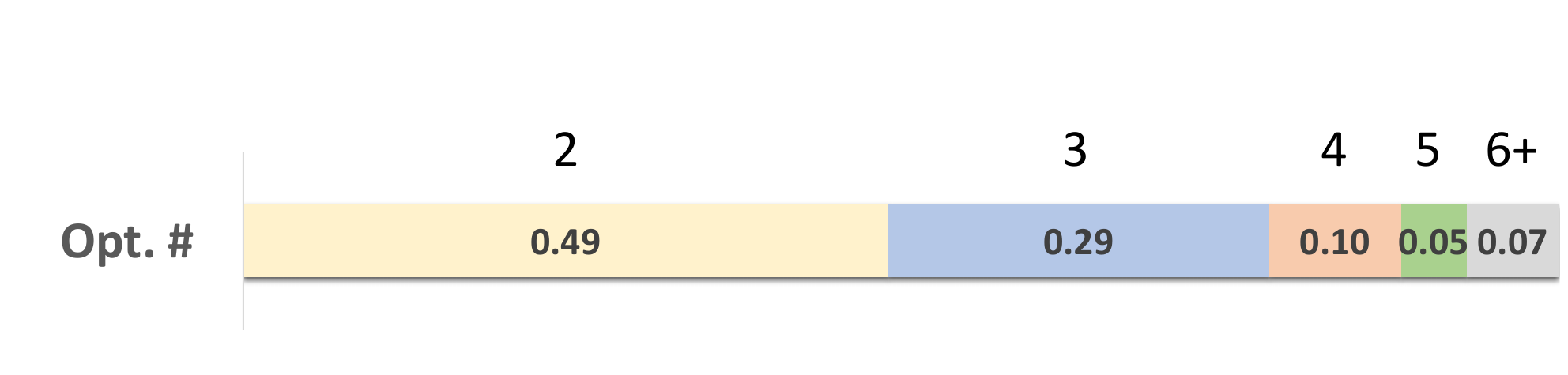}}
\caption{
Number of options distribution in \textsc{CAmbigNQ}.
}
\label{fig:opt_num}
\end{figure*}

Table~\ref{tab:category_statistic} provides an overview of the most frequently used categories within each split of the dataset. It is evident that the top five categories consistently appear in all three sets, although their specific order may vary.
This suggests that the dataset was well split into three parts.
Moreover, Figure~\ref{fig:category_distribution} illustrates the top 50 categories from the entire dataset, providing a broader perspective.
Overall, we have 593 distinct categories, with 412 of these categories occurring only once. 
This observation aligns with the previously discussed Inter-Annotator Agreement in Section~\ref{app:IAA}, where variations in specificity among annotators contribute to the presence of unique categories.
For instance, examples such as ``Jurrassic world" vs ``movie" or ``Will Turner" vs ``character" illustrate this inherent variability.
This existence of multiple representations for single categories or options is inherent to our task, and it is considered a natural occurrence.
The set of categories can be expanded as needed.
Additionally, we provide the number of options in our dataset in Figure~\ref{fig:opt_num}.
Since our dataset is built upon the \textsc{AmbigNQ}, the distribution of options is comparable to the previous study. 
% It is noteworthy that a significant majority, accounting for approximately 70\% instances in our dataset, demonstrate ambiguity with 2 to 3 distinct interpretations, as illustrated in Figure~\ref{fig:opt_num}.

\section{Details of human preference test}
\label{app:human_evaluation}
\subsection{Details of test setup}
We use Amazon Mechanical Turk\footnote{\url{https://www.mturk.com/}} for the human preference test.
To ensure the quality of responses, we restricted the workers whose nations are the US, CA, AU, NZ, UK and whose HIT minimum hits are over 5,000, and HIT rates are higher than 97\%.
Additionally, we enforced a requirement for annotators to provide at least one sentence explaining the reason for their choices.
Instances where annotators failed to provide a reason, provided a reason consisting of few words or presented a reason irrelevant to our task were deemed as ``rejected'' cases.
Annotators were informed of this rule and compensation for the MTurk workers was set at more than \$10 per hour.

\begin{figure*}[]
        \centering
        \begin{subfigure}[b]{0.9\textwidth}
            \centering
            \includegraphics[width=\textwidth]{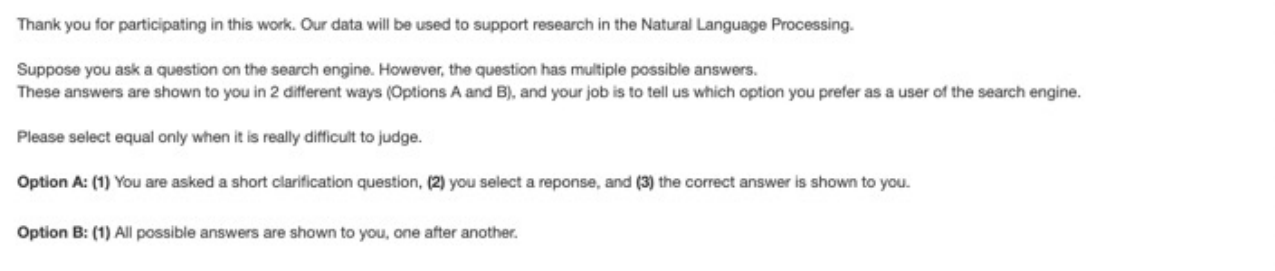}
            \caption[Network2]%
            {{\small Instructions given to MTURK workers.}}    
            \label{fig:mturk1}
        \end{subfigure}
        \hfill
        % \begin{subfigure}[b]{0.9\textwidth}   
        %     \centering 
        %     \includegraphics[width=\textwidth]{FIGURES/mturk2.pdf}
        %     \caption[]%
        %     {{\small MTURK interface page example 1}}    
        %     \label{fig:mturk2}
        % \end{subfigure}
        % \hfill
        % \begin{subfigure}[b]{0.9\textwidth}   
        %     \centering 
        %     \includegraphics[width=\textwidth]{FIGURES/mturk3.pdf}
        %     \caption[]%
        %     {{\small MTURK interface page example 2}}    
        %     \label{fig:mturk3}
        % \end{subfigure}
        % \hfill
        \begin{subfigure}[b]{0.9\textwidth}   
            \centering 
            \includegraphics[width=\textwidth]{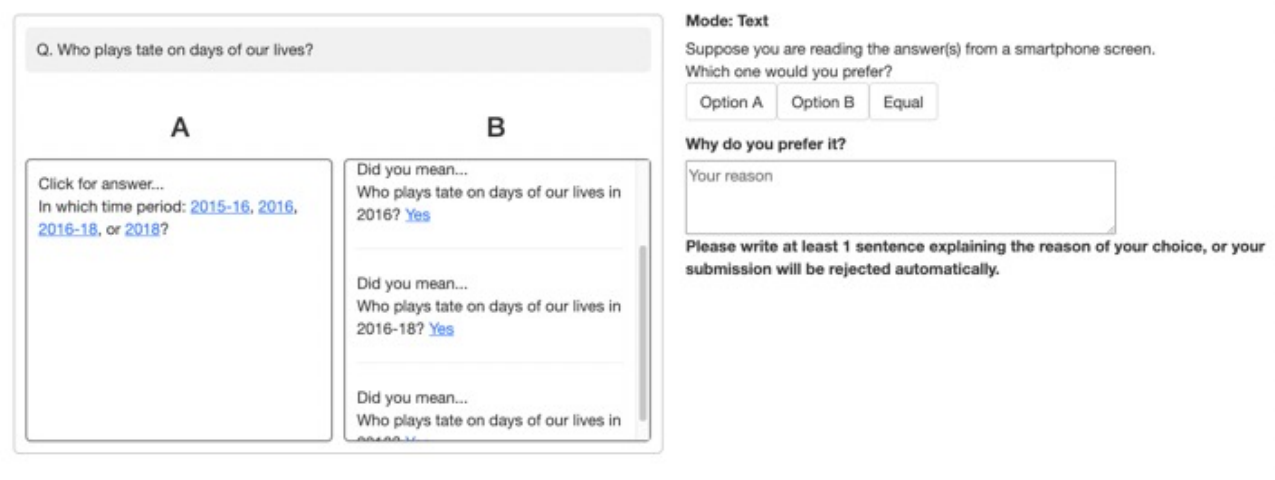}
            \caption[]%
            {{\small MTURK interface page example.}}    
            \label{fig:mturk4}
        \end{subfigure}
        \caption[ The average and standard deviation of critical parameters ]
        {Interface for preference test for MTURK workers.} 
        \label{fig:mturk}
    \end{figure*}

In order to investigate the potential correlation between the number of interpretations (i.e., the number of DQs) and user preference, we proceeded by partitioning our development dataset into five distinct groups based on the number of interpretations. These groups were categorized as data with two, three, four, five, and more than five interpretations. Subsequently, a single example was randomly selected from each group, resulting in the formation of one batch comprising five instances. Each batch was then assigned to an annotator for annotation, with a total of 20 batches being processed in this manner. 
An example of annotation interfaces is shown in Figure~\ref{fig:mturk}.

\subsection{Detailed analysis on test result}
\begin{figure*}
        \centering
        \begin{subfigure}[b]{0.475\textwidth}
            \centering
            \includegraphics[width=\textwidth]{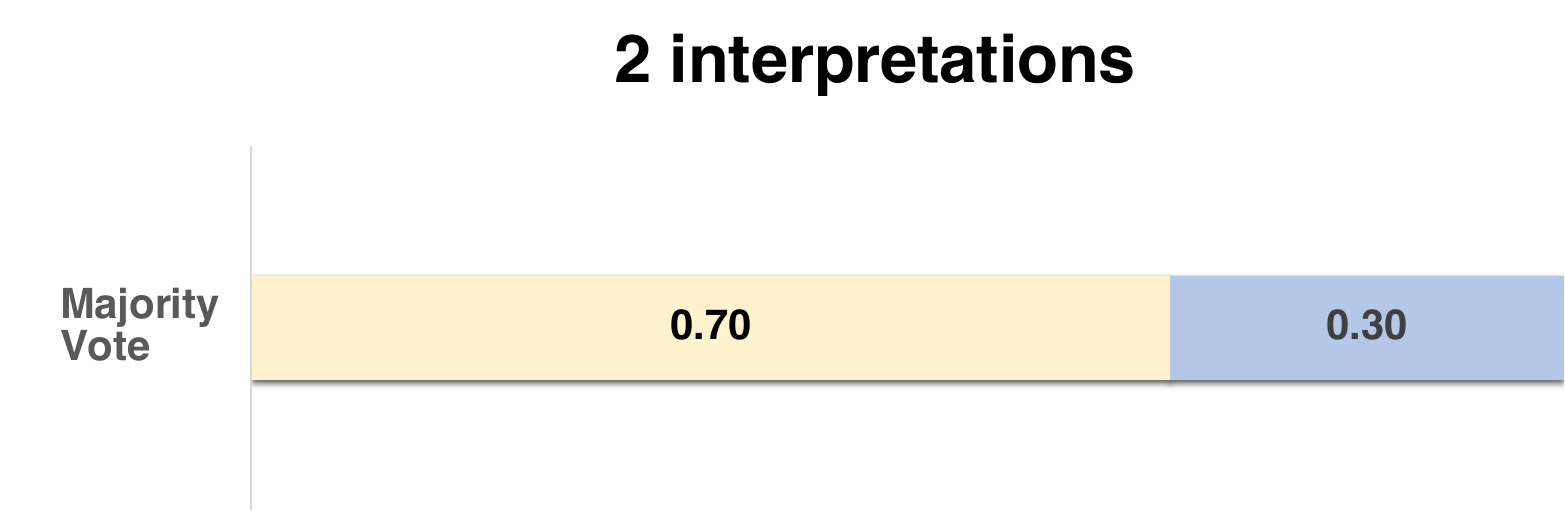}
            \caption[Network2]%
            {{\small Preference test result of two interpretations examples.}}    
            \label{fig:preference1}
        \end{subfigure}
        \hfill
        \begin{subfigure}[b]{0.475\textwidth}  
            \centering 
            \includegraphics[width=\textwidth]{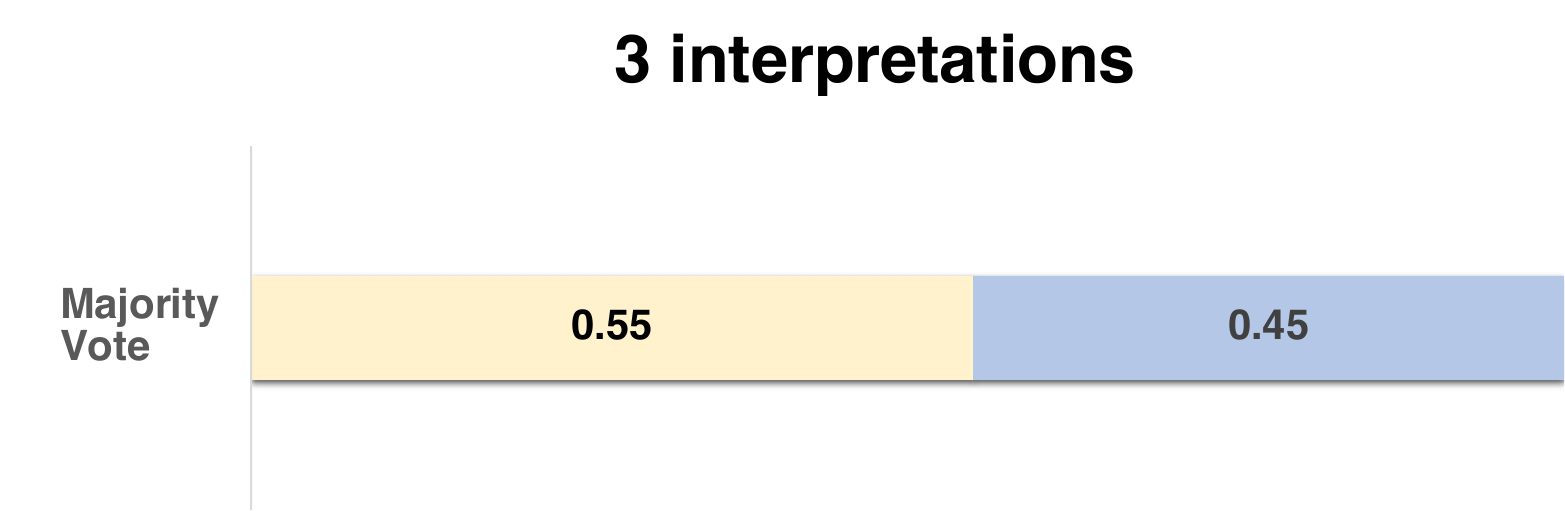}
            \caption[]%
            {{\small Preference test result of three interpretations examples.}}    
            \label{fig:preference2}
        \end{subfigure}
        \vskip\baselineskip
        \begin{subfigure}[b]{0.475\textwidth}   
            \centering 
            \includegraphics[width=\textwidth]{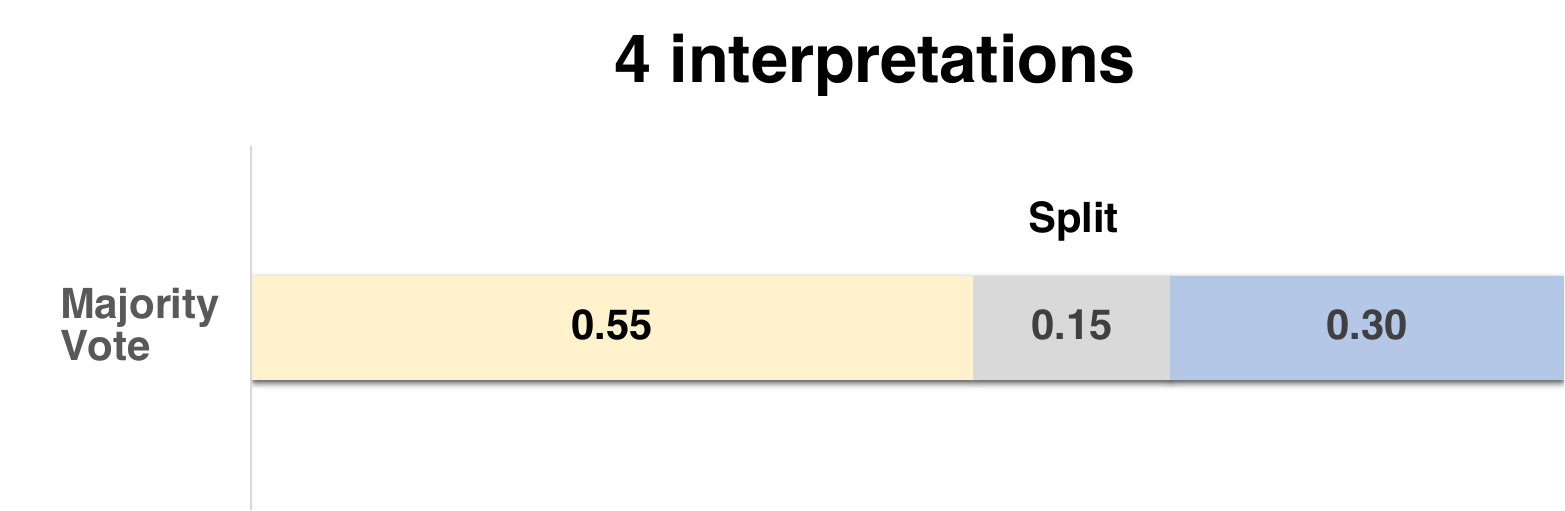}
            \caption[]%
            {{\small Preference test result of four interpretations examples.}}    
            \label{fig:preference3}
        \end{subfigure}
        \hfill
        \begin{subfigure}[b]{0.475\textwidth}   
            \centering 
            \includegraphics[width=\textwidth]{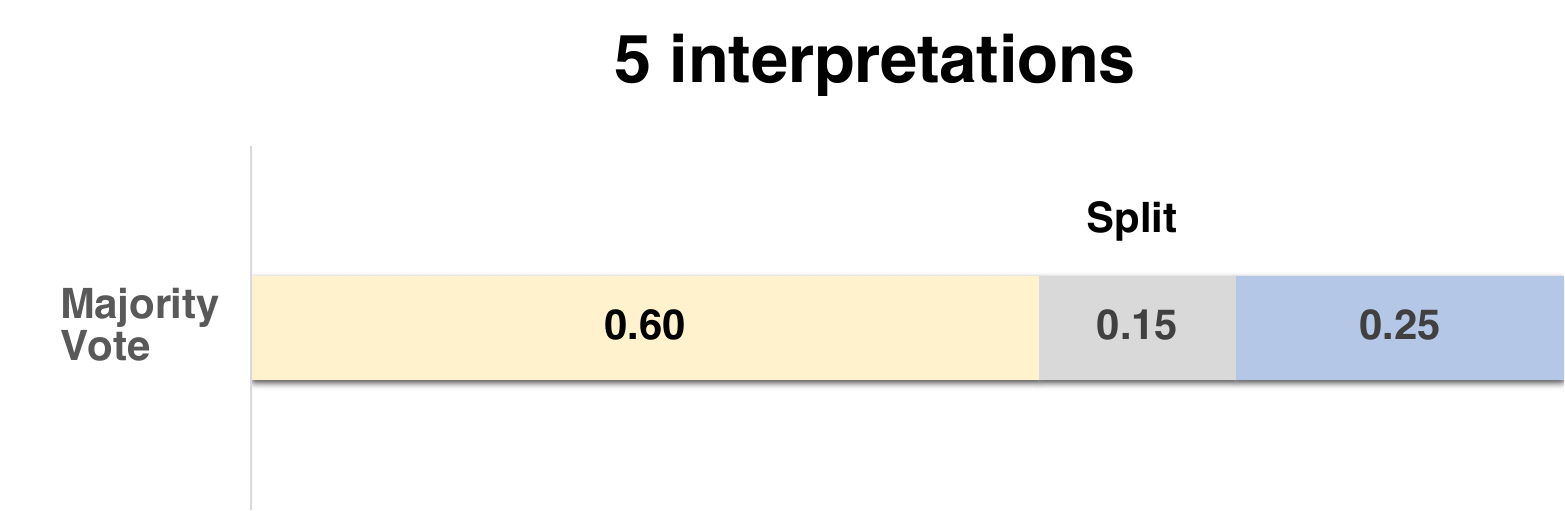}
            \caption[]%
            {{Preference test result of five interpretations examples.}}    
            \label{fig:preference4}
        \end{subfigure}
        \vskip\baselineskip
        \begin{subfigure}[b]{0.475\textwidth}  
            \centering 
            \includegraphics[width=\textwidth]{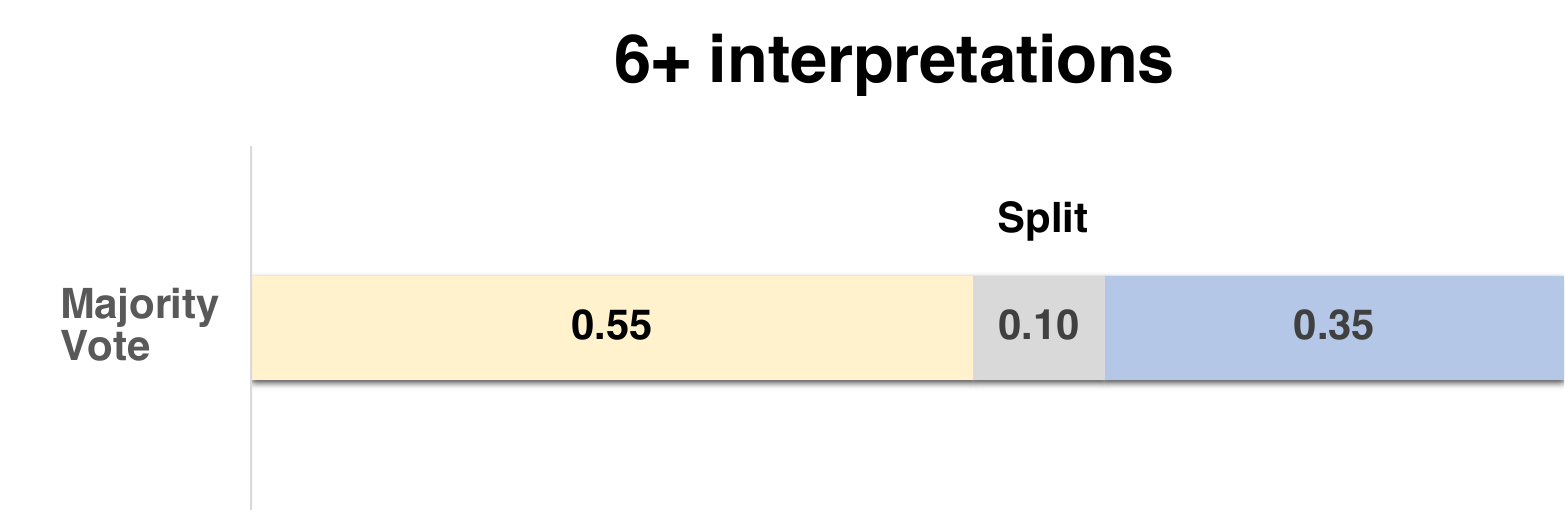}
            \caption[]%
            {{\small Preference test result of examples with more than five interpretations.}}    
            \label{fig:preference5}
        \end{subfigure}
        \caption[ The average and standard deviation of critical parameters ]
        {The preference test results for each group with a different number of interpretations.} 
        \label{fig:preferences}
    \end{figure*}

Figure~\ref{fig:preferences} provides the results obtained for varying numbers of interpretations.
The percentages of annotators favoring ``CQ'' remain relatively stable across different numbers of interpretations.

\section{Training Details}
\label{app:train_detail}
\paragraph{Training Detail of Ambiguity Detection}
The Ambiguity Detection task utilized a combined dataset consisting of 9,996, 977, and 977 instances in the train, validation, and test sets, respectively.
For the model, the BERT-base-cased model\footnote{\url{https://huggingface.co/bert-base-cased}} was finetuned with \textit{batch\_size} 16, \textit{accumulation\_step} 1, \textit{learning rate} \textit{2e-5}, and \textit{early\_stop\_patience} 1.
We use released checkpoint for pretrained \textsc{SpanSeqGen} model\footnote{\url{https://nlp.cs.washington.edu/ambigqa/models/ambignq-bart-large-12-0.zip}}.
We used one A5000 GPU for finetuning and it took approximately 4 hours. 
The training epochs were determined according to the validation performance based on accuracy.

\paragraph{Training Detail of Clarification Question Generation}
The two BART$_\text{large}$\footnote{\url{https://huggingface.co/facebook/bart-large}}\footnote{the total length of input was truncated to 1,024 tokens due to the maximum input length of the model} were finetuned on our \textsc{CambigNQ} with the training/validation/test split as described in Table~\ref{tab:data_statistics}.
Both models share the same hyperparameter during finetuning, which are \textit{batch\_size} 10, \textit{accumulation\_step} 2, \textit{learning rate} \textit{1e-5}, and \textit{early\_stop\_patience} 10.
The training epochs were determined according to the validation performance based on the BLEU score of the whole CQ.
We used one A6000 GPU for both finetuning and it took approximately 2 hours for \textit{No Answers} case and 4 hours for \textit{Predicted Answers} and \textit{Ground Truth Answers} cases.

\paragraph{Training Detail of Clarification-Based Question Answering}
For Clarification-Based Question Answering, the NQ-pretrained BART model\footnote{\url{https://nlp.cs.washington.edu/ambigqa/models/nq-bart-large-24-0.zip}} was finetuned with \textit{batch\_size} 10, \textit{accumulation\_step} 2, \textit{learning rate} \textit{1e-5}, and \textit{early\_stop\_patience} 10.
We used one A6000 GPU for finetuning and it took approximately 2 hours.
% While the percentage of annotators favoring CQ remains consistent at 59\% in both overall and majority vote, there is decrease in the percentage of annotators favoring DQ, decreased from 37\% to 33\%.
% This indicates that many examples where a single annotator opted for ``DQ" resulted in a split decision, highlighting the dominance of ``CQ" as the preferred choice.
% It is worth mentioning that the percentages of instances with a ``Split'' decision tend to increase when dealing with a higher number of interpretations.
% This observation suggests that the preference becomes more subjective in such cases, potentially due to the increased ambiguity and diverse perspectives.
\begin{table}[]
\centering
\resizebox{\columnwidth}{!}{%
\begin{tabular}{lcccc}
\hline
Reader Model &
  \multicolumn{1}{c}{Pre.} &
  \multicolumn{1}{c}{Rec.} &
  \multicolumn{1}{c}{F1} &
  \multicolumn{1}{c}{Acc.} \\ \hline
CQ finetuned BART &
  \textbf{58.0} &
  53.8 &
  \textbf{55.8} &
  35.8 \\
InstructGPT &
  7.4 &
  \textbf{60.0} &
  13.1 &
  \textbf{43.2} \\ \hline
\end{tabular}%
}
\caption{Evaluation results for the Clarification-based QA task employing two different reader models. Both cases utilize the truth CQs to clarify the AQ. 
The Acc. represents \textit{accuracy} which evaluates whether the model's response includes any gold answer.}
\label{tab:llm1}
\end{table}

\section{Inference Employing  Large Language Models}
Our primary experiments, which leveraged the BART-large models as our baselines, demonstrated suboptimal performance across different settings. 
To evaluate the efficacy of recent Large Language Models (LLMs) in our task, we designed additional experiments incorporating LLMs within our framework.
In these supplementary experiments, we only consider the most ideal case from Section~\ref{sec:cbqa_exp} where Ground Truth CQs are available.
\paragraph{Setup}
We leveraged two distinct variations of InstructGPT~\cite{Ouyang2022TrainingLM}, provided by OpenAI (namely, text-davinci-003, gpt-3.5-turbo), for our additional studies.
Initially, we employed the text-davinci-003 model (InstructGPT) as the reader model in Section~\ref{sec:cbqa_exp}, replacing the previously used BART-large models.
Subsequently, we reformulated our task as an interactive dialogue between the user and the QA models, comprising the following sequence: 1) User asking AQ, 2) Model offering CQs, 3) User selecting an option, and 4) Model generating corresponding answer for a given option.
Within this conversational framework, we utilized the gpt-3.5-turbo model (ChatGPT) and conducted inference under two settings: zero-shot and four-shot.

We evaluate both models using the conventional metrics of precision, recall, and F1 score.
Furthermore, due to the fact that both models generate responses at the sentence-level, quantifying the number of unique answers is challenging. 
Following ~\citet{liu2023lost} and ~\citet{mallen2022not}, we adopt \textit{accuracy}, which evaluates whether the prediction includes any gold answer.

\paragraph{Results and Analysis}
The evaluation results of clarification-based QA, utilizing InstructGPT as the reader model, and our task's reformulation within an interactive dialogue framework with ChatGPT, are presented in Table~\ref{tab:llm1} and Table~\ref{tab:llm2}, respectively.
It is noteworthy that the low precision in both results from InstructGPT, ChatGPT in zero-shot configuration, and ChatGPT in four-shot configuration can be attributed to the model's tendency to generate answers at the sentence level. These responses average 27.3 words, 25.4 words, and 19.9 words respectively. 
In contrast, the gold answers are more concise, averaging 2.6 words, leading to the observed low precision scores.

Utilizing InstructGPT as a reader model showed improved performance compared to our baseline, which uses CQ fine-tuned BART as a reader model.
Additionally, reformulating our task as an interactive dialogue and incorporating ChatGPT shows improved recall and accuracy.
However, it is evident that there is substantial potential for further enhancement which underscores both the challenging nature of our tasks and the need for further research.

% The evaluation results, derived from reformulating our task as an interactive dialogue framework using ChatGPT, are presented in Table~\ref{tab:llm2}.
% We found that ChatGPT struggled to generate distinct answers in both zero-shot and few-shot scenarios which again showcases the challenging nature of our tasks and the need for further research. 
% Similar to experiments conducted by using InstructGPT, ChatGPT yields low precision due to the sentence level generation, averaging 25.4 words in the zero-shot and 19.9 words in the four-shot scenario.

\begin{table}[]
\centering
\resizebox{\columnwidth}{!}{%
\begin{tabular}{lcccc}
\hline
ChatGPT &
  \multicolumn{1}{c}{Pre.} &
  \multicolumn{1}{c}{Rec.} &
  \multicolumn{1}{c}{F1} &
  \multicolumn{1}{c}{Accuracy} \\ \hline
Zero-shot &
  8.0 &
  \textbf{64.5} &
  14.3 &
  \textbf{50.8} \\
Four-shot &
  \textbf{11.3} &
  64.0 &
  \textbf{19.2} &
  49.9 \\ \hline
\end{tabular}%
}
\caption{Evaluation results of conversational setting employing ChatGPT. ChatGPT receives the input framed as an interactive dialogue between the user and the model, outlined in the subsequent sequence: 1) User asking AQ, 2) Model offering ground truth CQs, 3) User selecting an option, and 4) Model generating the corresponding answer for a selected option. Zero-shot and Four-shot denote the number of examples presented to the model within the prompt.}
\label{tab:llm2}
\end{table}

\begin{table*}[]
\centering
\small
\begin{tabular}{l|ll}
\cline{1-2}
Instruction &
  \begin{tabular}[c]{@{}l@{}}Generate the clarifying question for an   ambiguous question that gives options for corresponding \\ disambiguated   question.\end{tabular} &
   \\ \cline{1-2}
Example\_1 &
  \begin{tabular}[c]{@{}l@{}}ambiguous question: Why did the   st louis cardinals move to arizona? \\      disambiguated question 1:  what   ability caused the st louis cardinals move to arizona? \\      disambiguated question 2:  what   physical issue caused the st louis cardinals move to arizona?\\      disambiguated question 3: what fan issue caused the st louis cardinals move   to arizona?\\      clarifying question: Which type of reason: Ability, physical issue, or fan   issue?\end{tabular} &
   \\ \cline{1-2}
Example\_2 &
  \begin{tabular}[c]{@{}l@{}}ambiguous question: Who is the   current chairman of african union commission? \\      disambiguated question 1: who is the 4th chairman of african union   commission? \\      disambiguated question 2: who is the 3rd chairman of african union   commission? \\      disambiguated question 3: who is the 2nd chairman of african union   commission?\\      clarifying question: Which chairman: 4th, 3rd, or 2nd?\end{tabular} &
   \\ \cline{1-2}
Example\_3 &
  \begin{tabular}[c]{@{}l@{}}ambiguous question: Who won the   final hoh big brother 20? \\      disambiguated question 1: who won the final hoh in the american reality   show big brother 20? \\      disambiguated question 2: who won the final vote in the british reality   show celebrity big brother 20? \\      clarifying question: Which version: the american reality show, or the   british reality show celebrity?\end{tabular} &
   \\ \cline{1-2}
Example\_4 &
  \begin{tabular}[c]{@{}l@{}}ambiguous question: How long do   contestants get to answer on jeopardy? \\      disambiguated question 1: how long do contestants get to answer a typical   question on jeopardy? \\      disambiguated question 2: how long do contestants get to answer a final   jeopardy question on jeopardy? \\      disambiguated question 3: how long do contestants get to answer on jeopardy   's online test? \\      disambiguated question 4: how long do contestants have to answer during the   first two rounds of jeopardy? \\      clarifying question: For which type of questions: a typical question, a   final jeopardy question, jeopardy's\\ online test, or during the first two   rounds of jeopard?\end{tabular} &
   \\ \cline{1-2}
Example\_5 &
  \begin{tabular}[c]{@{}l@{}}ambiguous question: Who is the   longest serving manager in the premier league? \\      disambiguated question 1: who is the longest serving manager in the premier   league of all time in terms \\of time? \\      disambiguated question 2: who is the longest serving manager in the premier   league of all time in terms \\of number of games? \\      clarifying question: In terms of what: time, or the number of games?\end{tabular} &
   \\ \cline{1-2}
Example\_6 &
  \begin{tabular}[c]{@{}l@{}}ambiguous question: Who sang the   original do you love me? \\      disambiguated question 1: who is the band that sang the original do you   love me in 1962? \\      disambiguated question 2: who is the singer that sang the original do you   love me in for the contours \\in 1962? \\      disambiguated question 3: who are the characters that sang the original do   you love me in the fiddler\\ on the roof? \\      disambiguated question 4: who are the singers that sang the original do you   love me in the 1971 fiddler\\ on the roof film? \\      clarifying question: Which one: the band in 1962, the singer in the   contours in 1962, the characters in the \\fiddler on the roof, or the singer in   the 1971 fiddler on the roof film?\end{tabular} &
   \\ \cline{1-2}
\end{tabular}
\caption{The few-shot examples used for the prompt of the InstructGPT. These examples are concatenated with the instruction in certain order as mentioned in Section~\ref{sec:dataset}.}
\label{tab:fewshot}
\end{table*}

\begin{table*}[h]
\centering
\small
%\resizebox{\columnwidth}{!}{%
\begin{tabular}{ll}
\hline
Split & Categories (in the order of frequency)\\ \hline
Train & version, year, type, information, time    \\ 
Validation & version, type, time, year,  information   \\
Test       & version, type,  information, year, time \\
\hline
\end{tabular}%
%}
\caption{Most frequent categories in \textsc{CAmbigNQ}.}
\label{tab:category_statistic}
\end{table*}

\label{sec:appendix}
\end{document}